\documentclass[11pt]{article}

\usepackage{fullpage}

\usepackage[T1]{fontenc}
\usepackage[utf8]{inputenc}
\usepackage{lmodern}
\usepackage{microtype}

\usepackage{amsmath,mathtools}
\usepackage{amsfonts,amssymb}
\usepackage{amsthm,thmtools}
\usepackage{bm}
\usepackage{bbm}
\usepackage{accents}

\usepackage{graphicx}
\usepackage{xcolor}
\usepackage{booktabs}
\usepackage{array}
\usepackage{tabularx}
\usepackage{threeparttable}
\usepackage{caption}
\usepackage{subcaption}
\usepackage{makecell}
\usepackage{pifont}

\usepackage{enumitem}

\usepackage{algorithm}
\usepackage{algpseudocode}

\usepackage[numbers,compress]{natbib}
\usepackage[colorlinks=true,citecolor=blue,linkcolor=red,urlcolor=magenta]{hyperref}
\usepackage{cleveref}

\crefformat{equation}{(#2#1#3)}
\Crefformat{equation}{(#2#1#3)}
\crefrangeformat{equation}{(#3#1#4)--(#5#2#6)}
\Crefrangeformat{equation}{(#3#1#4)--(#5#2#6)}
\crefmultiformat{equation}{(#2#1#3)}{ and (#2#1#3)}{, (#2#1#3)}{, and (#2#1#3)}
\Crefmultiformat{equation}{(#2#1#3)}{ and (#2#1#3)}{, (#2#1#3)}{, and (#2#1#3)}

\theoremstyle{definition}
\newtheorem{theorem}{Theorem}[section]

\newtheorem{proposition}[theorem]{Proposition}

\newtheorem{definition}[theorem]{Definition}

\newtheorem{remark}[theorem]{Remark}

\newtheorem{example}[theorem]{Example}

\DeclarePairedDelimiterX{\cbrp}[1]{\lbrace}{\rbrace}{#1}
\DeclarePairedDelimiterX{\brp}[1]{(}{)}{#1}
\DeclarePairedDelimiterX{\sqbrp}[1]{[}{]}{#1}
\DeclarePairedDelimiterX{\normp}[1]{\lVert}{\rVert}{#1}
\DeclarePairedDelimiterX{\absp}[1]{\lvert}{\rvert}{#1}
\DeclarePairedDelimiterX{\ipp}[1]{\langle}{\rangle}{#1}

\DeclareRobustCommand{\term}[1]{%
  \ifmmode
    A_{\text{#1}}%
  \else
    term~\ensuremath{A_{\text{#1}}}%
  \fi
}

\DeclarePairedDelimiterX{\parens}[1]{(}{)}{#1}

\algnewcommand\algorithmicinit{\textbf{initialize}}
\algnewcommand\Init{\item[\algorithmicinit]}
\algnewcommand\algorithmicinput{\textbf{input}}
\algnewcommand\Input{\item[\algorithmicinput]}

\title{Chebyshev Center-Based Direction Selection for Multi-Objective Optimization and Training PINNs}

\author{
Hoyeol Yoon$^{1}$ \quad Seoungbin Bae$^{1}$ \quad Nam Ho-Nguyen$^{2}$ \quad Dabeen Lee$^{3}$\\[0.3em]
$^{1}$Department of Industrial \& Systems Engineering, KAIST\\
$^{2}$Discipline of Business Analytics, The University of Sydney\\
$^{3}$Department of Mathematical Sciences, Seoul National University\\
\texttt{9911hoho@kaist.ac.kr, sbbae31@kaist.ac.kr,}\\
\texttt{nam.ho-nguyen@sydney.edu.au, dabeenl@snu.ac.kr}
}

\date{}

\begin{document}
\maketitle

\begin{abstract}
Physics-informed neural networks (PINNs) are a promising approach for solving partial differential equations (PDEs). Their training, however, is often difficult because multiple loss terms induced by PDE residuals and boundary or initial conditions must be optimized simultaneously. To address this difficulty, existing approaches often construct update directions by explicitly enforcing particular desirable properties, such as scale robustness and simultaneous descent. While effective in many cases, such property-by-property designs can make it unclear which conditions are essential, what geometric principle determines the selected update direction, and how different methods are structurally related. In this work, we formulate update-direction selection for PINN training as a Chebyshev-center problem in the dual cone. The proposed formulation selects a normalized direction that maximizes the minimum distance to the cone facets. The resulting formulation admits an efficient dual problem in a much lower-dimensional space and yields a convergence guarantee in the nonconvex setting. It also recovers the key desirable properties targeted by existing approaches without imposing them separately;  rather, they follow from the single geometric criterion underlying the formulation. This makes the selected direction interpretable through a single geometric rule and provides a unified basis for systematically comparing related direction-selection methods. Experiments on several PINN benchmarks further demonstrate strong empirical performance of the proposed method.
\end{abstract}

\section{Introduction}
\label{sec:intro}

Partial differential equations (PDEs) are fundamental tools for modeling a wide range of phenomena in science and engineering, including fluid dynamics, heat transfer, and wave propagation. Since analytically solving complex PDEs is often intractable, a variety of numerical methods have been developed to approximate their solutions. Among recent learning-based approaches, physics-informed neural networks (PINNs) have emerged as a promising framework for solving PDEs~\cite{raissi2019physics, karniadakis2021physics}. In PINNs, a neural network is trained to approximate the PDE solution using objectives constructed from the governing equation together with boundary and initial conditions. This flexibility has enabled PINNs to be used for a broad class of forward and inverse problems. At the same time, however, their practical success depends critically on how effectively they can be trained~\cite{krishnapriyan2021characterizing, cuomo2022scientific}.

Training PINNs is challenging in practice, as their performance can be highly sensitive to data sampling, model design, and training dynamics~\cite{wu2023comprehensive,wang2021eigenvector,wang2021understanding}. These challenges have motivated a broad literature on sampling strategies~\cite{nabian2021efficient,tang2023pinns,gao2023failure}, network design~\cite{jagtap2020extended,cho2023separable,zhao2024pinnsformer}, and optimization~\cite{wang2022and,bischof2025multi}. In this work, we focus on the optimization aspect of PINN training, where multiple loss terms induced by PDE residuals, boundary conditions, and initial conditions must be optimized simultaneously. This multi-loss structure can produce imbalanced gradient magnitudes across objectives, while also creating conflicts among their preferred update directions~\cite{hwang2024dual,liu2025config}. As a result, the key optimization challenge is how to construct an update direction from multiple loss terms so that different physical constraints are reflected in a balanced manner.

Existing approaches to multi-loss optimization in PINNs can be broadly categorized into adaptive weighting methods and direction-selection methods. Adaptive weighting methods primarily aim to adjust the relative importance or the scales of different loss terms~\cite{wang2021understanding,wang2022and,bischof2025multi}, whereas direction-selection methods seek to construct an update direction from multiple loss-specific gradients~\cite{hwang2024dual,liu2025config,bu2026harmonized}. Many such direction-selection methods are motivated by desirable properties of the update direction, such as equalized progress across objectives, simultaneous descent, or robustness to gradient-scale differences. While these constructions are meaningful and often effective, designing directions by imposing such properties one by one can make it less clear which conditions are more essential, what geometric principle determines the selected direction, and how these methods are structurally related. 
These observations motivate a direction-selection formulation based on a single geometric criterion, rather than on separately imposed properties.

To this end, we 
formulate the direction-selection problem as a Chebyshev-center problem over the dual cone of the set of loss gradients. The dual cone provides natural candidates for update directions as it collects vectors that are descent directions for every loss function.
However, not all directions in the dual cone are equally desirable: a direction close to a facet may give only limited improvement to the corresponding objective. Motivated by this, based on the idea of Chebyshev centers, we select a normalized direction that serves as the center of the largest ball inscribed in the dual cone. Rather surprisingly, the single geometric formulation achieves several desirable properties all at once, which have been the individual targets of previous methods: scale robustness via normalized gradient comparisons, balanced treatment across objectives, and simultaneous descent. 
Moreover, the geometric interpretation, together with the corresponding primal and dual formulations, allows us to systematically compare related direction-selection methods.

In particular, we derive the dual of the Chebyshev center formulation, defined over a low-dimensional simplex, and show how to recover the corresponding primal update direction from the dual solution. This formulation also leads to a natural connection with Pareto stationarity, which enables a stopping criterion and a convergence analysis for the proposed algorithm. Then we compare our method with 
some representative methods, including DCGD~\cite{hwang2024dual}, HARMONIC~\cite{bu2026harmonized}, IMTL-G~\cite{liu2021towards}, ConFIG~\cite{liu2025config}, MGDA~\cite{sener2018multi} and GAPO~\cite{li2025gradient}, and the comparison shows that the proposed formulation can explain prior cone-based choices, recover desirable properties without imposing them separately, and relax restrictive equality-based formulations. Finally, we validate the proposed method through numerical experiments on a range of PINN problems.

\section{Preliminaries}
\label{sec:pre}

\paragraph{Physics-informed neural networks}
Physics-informed neural networks (PINNs)~\cite{raissi2019physics} approximate the solution \(u\) of a partial differential equation by a neural network \(u_\theta\) with parameters \(\theta\). Consider a PDE defined on a domain \(\Omega\), written as \(\mathcal{N}[u](x,t)=0\) for \((x,t)\in\Omega\), together with boundary and initial conditions \(\mathcal{B}[u](x,t)=0\) on \(\partial\Omega\) and \(\mathcal{I}[u](x,0)=0\) on \(\Omega\). The goal is to find \(\theta\) such that \(u_\theta\) approximately satisfies these physical constraints. To this end, PINNs sample collocation points from the interior of the domain, the boundary, and the initial slice, and define separate loss terms for each constraint, e.g.,
\(
\mathcal{L}_{\mathrm{r}}(\theta)
=
\frac{1}{N_r}\sum_{j=1}^{N_r}
\|\mathcal{N}[u_\theta](x_j^r,t_j^r)\|_2^2,
\)
\(
\mathcal{L}_{\mathrm{b}}(\theta)
=
\frac{1}{N_b}\sum_{j=1}^{N_b}
\|\mathcal{B}[u_\theta](x_j^b,t_j^b)\|_2^2,
\)
and
\(
\mathcal{L}_{\mathrm{i}}(\theta)
=
\frac{1}{N_i}\sum_{j=1}^{N_i}
\|\mathcal{I}[u_\theta](x_j^i,0)\|_2^2.
\)
Thus, PINN training naturally involves multiple loss terms associated with different physical constraints, and balancing them is often critical for stable and effective training. Related studies have been conducted to address this issue, which we review next.

\paragraph{Multi-objective perspectives on training PINNs}
To address the difficulty of balancing multiple PINN losses, prior work has explored several directions. One line focuses on adaptive weighting methods for PINNs, such as LRA~\cite{wang2021understanding}, NTK-based weighting~\cite{wang2022and}, and ReLoBRaLo~\cite{bischof2025multi}. 
Another line builds on classical multicriteria optimization, where a dual cone-based view of simultaneous descent was implicit in the characterization of descent directions~\cite{fliege2000steepest}.
Related to this classical viewpoint, gradient-based multi-objective and multi-task learning methods, including MGDA~\cite{sener2018multi}, PCGrad~\cite{yu2020gradient}, CAGrad~\cite{liu2021conflict}, and IMTL-G~\cite{liu2021towards}, combine or modify gradients to reduce conflict among objectives. More recently, gradient-based multi-objective methods have also been developed for constrained reinforcement learning, preference-based MOO/MORL, and multi-objective LLM alignment, including GCPO~\cite{zhou2023gradient}, PMGDA~\cite{zhang2025pmgda}, GAPO~\cite{li2025gradient}, and RACO~\cite{chen2026reward}. In the PINN setting, DCGD~\cite{hwang2024dual}, ConFIG~\cite{liu2025config}, and HARMONIC~\cite{bu2026harmonized} are closely related methods that construct update directions from multiple loss gradients. Other PINN-specific optimization strategies include the parameter-wise adaptive optimizer MultiAdam~\cite{yao2023multiadam} and the minimax-based method BGDA~\cite{bylinkin2026enhancing}. In this paper, we focus on gradient-based methods for determining update directions 
and compare methods that are closely related to ours in Section~\ref{sec:compare}.

\paragraph{Chebyshev centers of polyhedral sets}
The Chebyshev center is a classical notion for defining the center of a constrained geometric set (see~\cite{boyd2004convex}). Given a norm \(\|\cdot\|\), the depth of \(x \in C\) for a bounded set \(C \subset \mathbb{R}^n\) with nonempty interior is \(\inf_{y\in \mathbb{R}^n\setminus C}\|x-y\|\), namely the radius of the largest norm ball centered at \(x\) and contained in \(C\). A Chebyshev center is a point of maximum depth, or equivalently the center of a largest inscribed norm ball. For a polyhedron \(C=\{x\mid a_i^\top x\le b_i,\ i=1,\ldots,m\}\), requiring a norm ball of radius \(R\) centered at \(x\) to lie in \(C\) is equivalent to \(a_i^\top x+R\|a_i\|_*\le b_i\) for all \(i\), where \(\|\cdot\|_*\) denotes the dual norm of \(\|\cdot\|\). Thus, the Chebyshev center gives a notion of centrality for polytopes. At the same time, it naturally extends to polyhedral cones, explained in Section~\ref{sec:chebyshev}.

\section{Methodology}
\label{sec:method}

\subsection{
Chebyshev center of the dual cone for direction selection}\label{sec:chebyshev}
We consider a multi-objective optimization problem with \(m\) loss functions \(\{\mathcal{L}_i\}_{i=1}^m\).
At iteration \(t\), let \(g_i := \nabla \mathcal{L}_i(\theta_t)\) denote the gradient of $\mathcal{L}_i$.
Under standard gradient descent, the rate of change in 
\(\mathcal{L}_i\) along a direction \(d\) is determined by \(g_i^\top d\); in particular, if \(g_i^\top d < 0\), $d$ is not a descent direction for \(\mathcal{L}_i\).
Therefore, depending on the choice of \(d\), some losses may decrease while others increase. To avoid such conflicts, we consider directions satisfying \(g_i^\top d \ge 0\) for all \(i=1,\dots,m\).
This simultaneous descent condition can be naturally expressed in terms of the dual cone.

\begin{definition}[Dual cone]\label{def:dual_cone}
Let \(\mathbf{K} \subseteq \mathbb{R}^n\) be a cone. Then, the dual cone of \(\mathbf{K}\) is defined as
\[\mathbf{K}^* = \{\, y \in \mathbb{R}^n \mid x^\top y \geq 0, \; \forall x \in \mathbf{K}\,\}.\]
\end{definition}
Let \(\mathbf{K}_t := \operatorname{cone}(g_1,\dots,g_m)\) be the cone generated by the gradients at iteration \(t\), and let \(\mathbf{K}_t^*\) denote its dual cone. If \(d \in \mathbf{K}_t^*\), then \(x^\top d \ge 0\) for all \(x \in \mathbf{K}_t\). In particular, \(g_i^\top d \ge 0\) for all \(i=1,\dots,m\).
Thus, choosing the update direction \(d\) from \(\mathbf{K}_t^*\) guarantees a descent direction for all losses. As the dual cone may contain infinitely many feasible directions, an additional criterion is needed to choose an update direction. Inspired by this, in this work, we develop a centrality notion within the dual cone as the selection criterion, favoring directions that lie away from the cone boundaries.


Note that hyperplane \(g_i^\top d = 0\) defines a face 
of the cone. Since \(g_i^\top d\) determines the rate of change in $\mathcal{L}_i$, choosing a direction near this face yields a small rate of decrease for the corresponding loss. This motivates a centrality criterion that favors directions away from the cone boundary. 
To formalize this, we use the Chebyshev center, a geometric notion that defines a center of a polyhedral set.

To elaborate, we consider a pair of conjugate \(\ell_p/\ell_q\) norms with \(1\leq p\leq\infty\) and \(q=p/(p-1)\). 
Let \(\hat g_i^{(p)} := {g_i}/{\|g_i\|_p}\) denote the gradient normalized with respect to the \(\ell_p\) norm. Throughout the paper, we assume that \(\|g_i\|_p>0\) for all \(i\), so that each normalized gradient is well defined. Since \(\hat g_i^{(p)}\) is a positive rescaling of \(g_i\), the hyperplane \(g_i^\top v = 0\) is equivalently written as \((\hat g_i^{(p)})^\top v=0\). 
Then (the absolute value of) the distance from \(v\) to this hyperplane is
\({g_i^\top v}/{\|g_i\|_p} = (\hat g_i^{(p)})^\top v\). Here, the Chebyshev-center criterion seeks a center \(v\) that maximizes the smallest of the distances to the hyperplanes. Because the dual cone is scale-invariant, this criterion requires normalization of \(v\); otherwise, the 
distances can grow trivially along the same ray. 
Since our interest is in the update direction rather than its magnitude, we fix the scale by imposing \(\|v\|_q \le 1\).
Note that finding a Chebyshev center boils down to computing the maximum admissible radius of an $\ell_q$ ball contained in the dual cone. Hence, we deduce the following formulation.
\begin{equation}\label{primal}
\tag{P}
\max_{v,\,r}\quad  r
\quad 
\text{s.t.}\quad
(\hat{g}_i^{(p)})^\top v \ge r,\; i = 1, \dots, m,\quad 
\|v\|_q \le 1.
\end{equation}

Note that \((v,r)=(0,0)\) is always feasible for Problem \eqref{primal}, and hence the optimal value satisfies \(r^\star \ge 0\). Therefore, any optimal solution \(v^\star\) satisfies \((\hat g_i^{(p)})^\top v^\star \ge 0\) for all \(i=1,\dots,m\). Since \(\hat g_i^{(p)}\) is a positive rescaling of \(g_i\), it follows that \(g_i^\top v^\star \ge 0\) for all \(i\), and hence \(v^\star \in \mathbf{K}_t^*\).

\begin{figure}[t]
    \centering
    \includegraphics[width=0.8\linewidth]{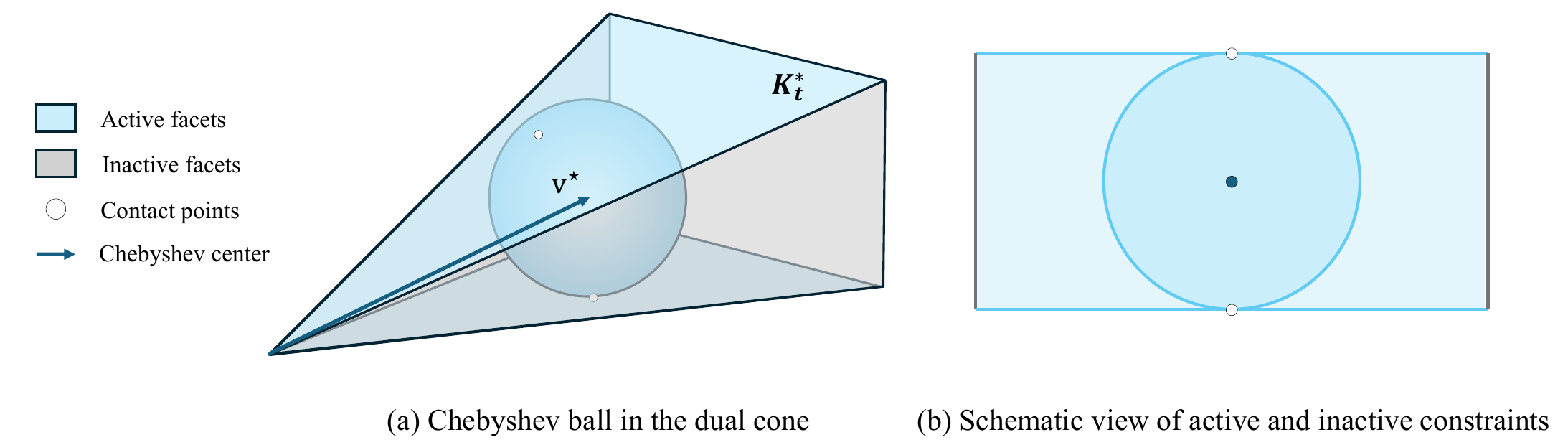}
    \caption{
    Geometric illustration of the proposed direction-selection rule, shown in the Euclidean case \(p=q=2\) for visual clarity.
    (a) The update direction \(v^\star\) is chosen from the dual cone \(\mathbf{K}_t^*\) as the center of an admissible Chebyshev ball.
    (b) A schematic cross-sectional view shows that only active constraints are tangent to the ball, while inactive constraints remain separated.
    The same active/inactive interpretation applies under general $\ell_q$ balls.
    }
    \label{fig:chebyshev_ball}
\end{figure}

The radius \(r\) is determined by the closest facets to the center \(v\); hence, only the closest facets are active at the optimum. 
Other ones may remain strictly farther away, as illustrated in Figure~\ref{fig:chebyshev_ball}. 
Therefore, the formulation does not enforce that the distances to the hyperplanes are equal, but only requires them to be at least \(r\). 
This captures the largest admissible inscribed $\ell_q$ ball centered at the selected direction.

\subsection{Dual formulation to get optimal solution efficiently}

We next derive the dual formulation of \eqref{primal} for efficient computation of the update direction.

\begin{proposition}\label{prop:dual_formulation}
The dual of \eqref{primal} is given by 
\begin{equation}\label{dual}
\tag{D}
\min_{\alpha} \left\| \sum_{i=1}^m \alpha_i \hat g_i^{(p)} \right\|_p 
\quad
\text{s.t.}\quad
\sum_{i=1}^m \alpha_i = 1,
\quad
\alpha_i \ge 0,\; i=1,\dots,m.
\end{equation}
Moreover, strong duality holds between~(P) and~(D).
\end{proposition}

By Proposition~\ref{prop:dual_formulation}, \eqref{primal} can be solved equivalently through its dual Problem~\eqref{dual}. This is advantageous in our setting because the primal variable \(v\) lies in the parameter space \(\mathbb{R}^n\), whose dimension \(n\) is typically very large in neural network training. In contrast, the dual variable \(\alpha\) lies in \(\mathbb{R}^m\), where \(m\) is the number of loss terms and is usually much smaller than \(n\). In such case, solving the dual is computationally more favorable than solving the primal problem directly, since it reduces the optimization to a simplex-constrained convex problem in a much lower-dimensional space.

Although \eqref{primal} can be solved efficiently through its dual formulation, our main interest is in the primal optimal direction \(v^\star\), which is used as the update direction. Solving the dual problem yields the optimal dual variable \(\alpha^\star\), and strong duality ensures agreement of the optimal values, but it does not directly provide the primal direction itself. The following proposition shows that, when \(r^\star>0\), the primal optimal direction can be recovered explicitly from the dual optimal solution.

\begin{proposition}\label{prop:recovery_direction}
Assume that \(1<p<\infty\). Let \( (v^\star, r^\star) \) be an optimal solution to the primal \eqref{primal}, and let \( \alpha^\star \) be an optimal solution to the dual \eqref{dual}. Define
$w^\star := \sum_{i=1}^m \alpha_i^\star \hat{g}_i^{(p)}$.
If \( r^\star > 0 \), then \(w^\star\neq 0\) and the primal optimal direction is recovered as
$v^\star
={
\operatorname{sgn}(w^\star)\odot |w^\star|^{p-1}
}/{
\|w^\star\|_p^{p-1}
}$,
where \(\operatorname{sgn}(\cdot)\), \(|\cdot|\), and the power are applied componentwise, and \(\odot\) denotes componentwise multiplication.
\end{proposition}

Note that when \(p=2\), the recovery formula in Proposition~\ref{prop:recovery_direction} reduces to the simple Euclidean normalization \(v^\star=\frac{w^\star}{\|w^\star\|_2}\). For \(1<p<\infty\), Proposition~\ref{prop:recovery_direction} shows that the primal update direction can be obtained directly from the dual optimal solution. Therefore, in practice, it is sufficient to solve \eqref{dual}, form \(w^\star = \sum_{i=1}^m \alpha_i^\star \hat g_i^{(p)}\), and recover \(v^\star\) using the formula in Proposition~\ref{prop:recovery_direction}. This provides an efficient way to compute the update direction without solving the primal problem explicitly.

\begin{remark}\label{remark:recovery}
The cases \(p=1\) and \(p=\infty\) are also well defined for both the primal and dual formulations. However, the closed-form map in Proposition~\ref{prop:recovery_direction} no longer directly recovers the primal direction from the dual solution, as the corresponding primal direction may not be uniquely determined. Related \(\ell_\infty\)-norm constraints also arise in classical multicriteria steepest-descent methods; in that setting, the resulting direction-selection problem can be formulated as a linear program~\cite{fliege2000steepest}. In Appendix~\ref{app:p1_pinf_recovery}, we spell out the corresponding  recovery step explicitly for our formulation.
\end{remark}

The same primal--dual framework also provides a natural stopping criterion for the method. To formalize this criterion, we use Pareto stationarity as the first-order notion of optimality in multi-objective optimization.

\begin{definition}[Pareto stationarity]\label{def:pareto_stat}
For the multi-objective problem
$\min_{\theta \in \mathbb{R}^n} \; \bigl(\mathcal{L}_1(\theta), \dots, \mathcal{L}_m(\theta)\bigr)$
where each \(\mathcal{L}_i\) is differentiable, a point \(\theta^\star \in \mathbb{R}^n\) is \emph{Pareto-stationary} if there exists \(\alpha \in \mathbb{R}_+^m\) such that
$\sum_{i=1}^m \alpha_i = 1$ and $\sum_{i=1}^m \alpha_i \nabla \mathcal{L}_i(\theta^\star) = 0$.
\end{definition}

\begin{proposition}\label{prop:r0_stationary}
Let \( (v^\star, r^\star) \) be an optimal solution to the primal \eqref{primal} at \(\theta\). If \( r^\star = 0 \), then \(\theta\) is Pareto-stationary.
\end{proposition}

Proposition~\ref{prop:r0_stationary} shows that the primal optimal value provides a natural stopping criterion for our method. In practice, however, requiring \(r^\star = 0\) exactly is too restrictive, so we use an approximate criterion instead. Specifically, we refer to a point \(\theta_t\) as \(\epsilon\)-Pareto-stationary if the corresponding primal optimal value satisfies \(r^\star_t \le \epsilon\) for a prescribed tolerance \(\epsilon > 0\). By strong duality, this can be checked through
\(\|\sum_{i=1}^m \alpha_{t,i}\hat g_i^{(p)}(\theta_t)\|_p,\)
which is computed in the algorithm. With this stopping criterion, our method can be summarized in Algorithm~\ref{alg:dual-fw}. Since the dual is a convex problem over a simplex, one may use the Frank--Wolfe algorithm, as for MGDA~\cite{sener2018multi}.
At the same time, in the \(\ell_2\) case with \(m\) is small, there exists a faster dual solver; see Appendix~\ref{app:algorithms}. In the final step, the recovered \(\ell_q\)-unit direction is scaled by the sum of its inner products with all loss gradients to form the final direction.

\begin{algorithm}[t]
\caption{Chebyshev Center-Based Direction Search via the Dual Formulation (DualChebyshev)}
\label{alg:dual-fw}
\begin{algorithmic}[1]
\Require Initial point \(\theta_0\), norm parameter \(p\in(1,\infty)\), step sizes \(\{\eta_t\}_{t=0}^{T-1}\), threshold \(\epsilon > 0\)
\For{\(t = 0,1,\dots,T-1\)}
    \State Compute \(\ell_p\)-normalized gradients \( \hat{g}_1^{(p)}(\theta_t), \dots, \hat{g}_m^{(p)}(\theta_t) \)
    \State Solve the dual \eqref{dual} to obtain \( \alpha_t \in \Delta^m \)
    \State Form
    $w_t \gets \sum_{i=1}^m \alpha_{t,i}\,\hat{g}_i^{(p)}(\theta_t)$
    \If{\( \|w_t\|_p \le \epsilon \)}
        \State \textbf{terminate} and return \(\theta_t\)
    \EndIf
    \State Recover the primal direction:
    $v_t \gets
    {
    \operatorname{sgn}(w_t)\odot |w_t|^{p-1}
    }/{
    \|w_t\|_p^{p-1}
    }$
    \State Set the final update direction with adaptive scalar:
    $d_t \gets (\sum_{i=1}^m g_i(\theta_t)^\top v_t)\,v_t$
    \State Update the iterate:
    $\theta_{t+1} \gets \theta_t - \eta_t d_t$
\EndFor
\State \Return \(\theta_T\)
\end{algorithmic}
\end{algorithm}

\begin{remark}\label{remark:algorithm}
For the cases \(p=1\) and \(p=\infty\), the implementation differs from Algorithm~\ref{alg:dual-fw}. 
The dual objective becomes nonsmooth, so the dual problem is better handled through an LP reformulation rather than the Frank--Wolfe step used above. Moreover, as discussed in Remark~\ref{remark:recovery}, these cases require an additional recovery step because the closed-form map in Proposition~\ref{prop:recovery_direction} does not directly apply.
Appendix~\ref{app:p1_pinf_dual} describes how to solve the dual problem in the \(p=1\) and \(p=\infty\) cases, and Appendix~\ref{app:p1_pinf_recovery} describes the corresponding recovery step.
\end{remark}

\subsection{Convergence analysis}

The following theorem provides a convergence guarantee for Algorithm~\ref{alg:dual-fw} in the nonconvex setting under a smoothness assumption on the total loss that is compatible with the chosen \(\ell_q\) geometry. Here, \(\theta^\star\) denotes a reference point satisfying \(\mathcal{L}(\theta^\star) \le \mathcal{L}(\theta_t)\) for all \(t\).

\begin{theorem}\label{thm:convergence}
Assume that the loss functions \(\mathcal{L}_1, \mathcal{L}_2, \dots, \mathcal{L}_m\) are differentiable and the total loss $\mathcal{L}(\theta) := \sum_{i=1}^m \mathcal{L}_i(\theta)$
is \(\beta\)-smooth with respect to the $\ell_q$ norm, i.e.,
\(\mathcal{L}(\theta') \le \mathcal{L}(\theta) +
\nabla \mathcal{L}(\theta)^\top(\theta'-\theta) +
\frac{\beta}{2}\|\theta'-\theta\|_q^2 \)
for all \(\theta,\theta'\). Let \(\{\theta_t\}_{t=0}^{T}\) be the sequence generated by Algorithm~\ref{alg:dual-fw}. Then, for a step size \(\eta = {1}/{\beta}\), the sequence either terminates at an \(\epsilon\)-Pareto-stationary point, or satisfies
\[
\min_{0 \le t \le T-1}\|\nabla \mathcal{L}(\theta_t)\|_p^2
\le
\frac{2\beta\bigl(\mathcal{L}(\theta_0)-\mathcal{L}(\theta^\star)\bigr)}{\epsilon^2T}.
\]
\end{theorem}

This implies that, unless Algorithm~\ref{alg:dual-fw} terminates earlier at an \(\epsilon\)-Pareto-stationary point, the best iterate among \(\{\theta_t\}_{t=0}^{T-1}\), in terms of the total-loss gradient norm, satisfies an \(O(1/\sqrt{T})\) bound.

\section{Comparative analysis}
\label{sec:compare}

In this section, we compare our method with several closely related direction-selection methods. 
Recall that ours selects a direction through the Chebyshev-center criterion over the dual cone, rather than by imposing individual properties. We will see that based on the primal and dual formulations for the Chebyshev center, we may  systematically analyze and compare the related methods.

\begin{figure}[t]
    \centering
    \includegraphics[width=0.95\linewidth]{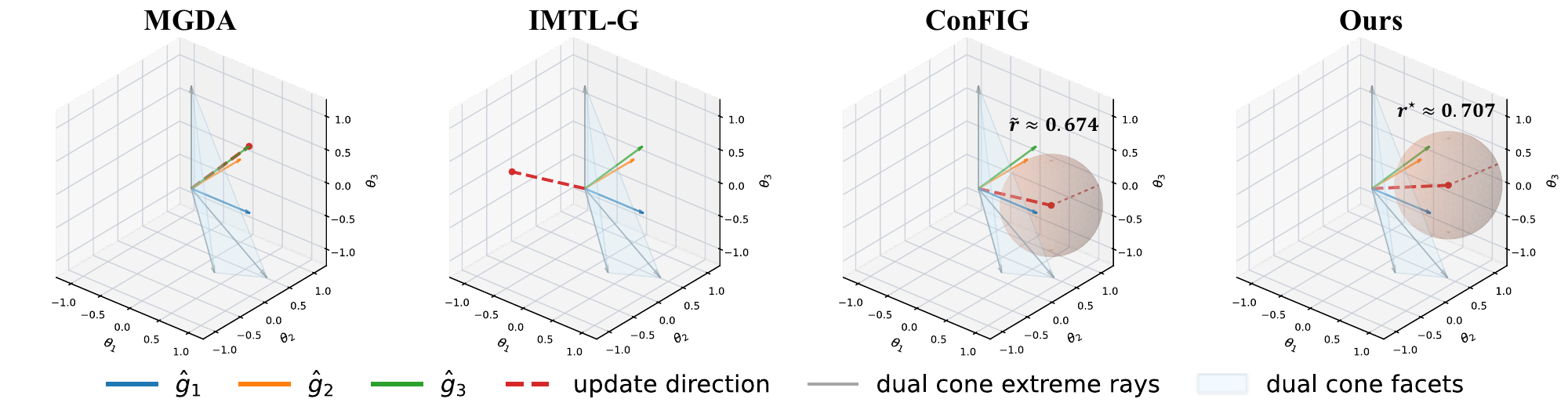}
    \caption{
    A three-objective  example illustrating geometric differences with some representative methods.
    }
    \label{fig:method_comparison}
\end{figure}

Since the related methods considered below are formulated in the Euclidean geometry, we focus in this section on the default case \(p=q=2\). Accordingly, we write \(\hat g_i := \hat g_i^{(2)}={g_i}/{\|g_i\|_2}\)
throughout this section. For ease of comparison, we express all methods in this Euclidean notation.
Figure~\ref{fig:method_comparison} gives a geometric overview of the four direction-selection methods compared in this section. The construction details for this illustration are provided in Appendix~\ref{app:comparison_toy}.

\subsection{Geometric comparison with cone-based methods}

Cone-based methods use conic geometry to construct update directions from multiple loss gradients. 
In this subsection, we compare our method with DCGD~\cite{hwang2024dual} and HARMONIC~\cite{bu2026harmonized}, focusing on how their constructions relate to our Chebyshev-center formulation.

    \paragraph{DCGD}
    Within the dual cone, three direction-selection variants are considered: Projection, Average, and Center~\cite{hwang2024dual}. 
    Among them, the Center variant uses the angle bisector and was reported to achieve the best empirical performance. In the two-objective nondegenerate case, the angle bisector coincides with the direction selected by our Chebyshev-center formulation. 
    Thus, for \(m=2\), the reported advantage of DCGD (Center) is consistent with the centrality criterion used in our method.
    
    For \(m \ge 3\), however, the notion of an angle bisector is no longer canonical, whereas our Chebyshev-center formulation may take any number of objectives.

\paragraph{HARMONIC}
HARMONIC~\cite{bu2026harmonized} constructs the update direction from the harmonized cone \(\mathbf{H}_t:=\mathbf{K}_t\cap \mathbf{K}_t^*\), where \(\mathbf{K}_t\) is the primal gradient cone and \(\mathbf{K}_t^*\) is its dual cone. Candidates are parameterized as \(d=G_t\lambda\), where \(G_t=[g_1,\dots,g_m]\) and \(\lambda\in\mathbb{R}^m_+\), so that \(d\in\mathbf{K}_t\).
Then the condition \(d\in\mathbf{K}_t^*\) is  imposed as
\(G_t^\top d=G_t^\top G_t\lambda\ge0_m\).
It then enumerates the extreme rays of the resulting coefficient-space cone via the Double Description method, maps them back through \(G_t\), and averages the normalized rays.
Thus, HARMONIC enforces membership in both \(\mathbf{K}_t\) and \(\mathbf{K}_t^*\) by construction.

In contrast, our method neither constructs \(\mathbf{H}_t\) nor enumerates extreme rays.
The Chebyshev-center problem \eqref{primal} directly selects a direction in \(\mathbf{K}_t^*\). Moreover, Proposition~\ref{prop:recovery_direction} gives \(v^\star = w^\star/\|w^\star\|_2\), where \(w^\star\) is a conic combination of the normalized gradients and hence lies in \(\mathbf{K}_t\). 
Therefore, the \(\mathbf{H}_t\) membership, explicitly enforced by HARMONIC, automatically holds in our method as an optimality consequence of the Chebyshev-center formulation.

\subsection{Comparison with equality-based methods based on the primal formulation}

We next compare our formulation with methods that determine the update direction through gradient--direction inner products.
In particular, IMTL-G~\cite{liu2021towards} and ConFIG~\cite{liu2025config} both aim to equalize these inner products across objectives, but they differ in whether the resulting direction is guaranteed to lie in the dual cone. 
Through \eqref{primal}, we view these methods as equality-based counterparts of our inequality-based Chebyshev-center formulation.

\paragraph{IMTL-G}
IMTL-G~\cite{liu2021towards} seeks a direction \(d=\sum_{i=1}^m \alpha_i g_i\), with \(\sum_{i=1}^m \alpha_i=1\), whose normalized-gradient inner products are equal across objectives: \(\hat g_1^\top d=\hat g_2^\top d=\cdots=\hat g_m^\top d.\)
Under this condition, IMTL-G yields a closed-form expression for the coefficients \(\alpha\). 
Specifically, letting \(\alpha=[\alpha_2,\ldots,\alpha_m]^\top\), \(U^\top=[\hat g_1-\hat g_2,\ldots,\hat g_1-\hat g_m]\), and \(G^\top=[g_1-g_2,\ldots,g_1-g_m]\), the resulting coefficients are given by \(\alpha=g_1^\top U(G^\top U)^{-1}\).

Thus, IMTL-G explicitly normalizes gradients and imposes equality of their inner products with the update direction.
However, as noted in ConFIG~\cite{liu2025config}, the common value can be negative; if \(\hat g_1^\top d=\cdots=\hat g_m^\top d=c<0\), then the resulting direction does not lie in \(\mathbf{K}_t^*\).

In contrast, in our formulation, normalization is not introduced as a separate designed property; it follows from measuring distances to the hyperplane of the dual cone.
Moreover, the update direction is selected within \(\mathbf{K}_t^*\), avoiding the negative inner product value described above.

\paragraph{ConFIG}
ConFIG~\cite{liu2025config} normalizes the gradients and sets their inner products with the update direction to the same positive value.
In our notation, it solves \(\hat G^\top w=\mathbf{1}_m\) by the pseudo-inverse,
\(w=(\hat G^\top)^+\mathbf{1}_m\), where \(\hat G=[\hat g_1,\ldots,\hat g_m]\), and uses the normalized direction \(v=w/\|w\|_2\).
A positive scalar rescaling is then applied to form the final update. In fact, one may equivalently formulate ConFIG as the following equality-constrained counterpart of \eqref{primal}:
\begin{equation}\label{primal_eq}
\tag{P$_{=}$}
\max_{v,\,r}\quad r \quad \text{s.t.} \quad \hat{g}_i^\top v = r, \; i = 1, \dots, m,\quad \|v\|_2 \le 1.
\end{equation}
By a change of variables, Problem \eqref{primal_eq} reduces to the pseudo-inverse minimum-norm system used by ConFIG. The detailed derivation is provided in Appendix~\ref{app:config_equiv}.

Moreover, the following theorem establishes another relationship between \eqref{primal} and \eqref{primal_eq}.

\begin{theorem}\label{thm:config_comparison}
Let \((v^\star,r^\star)\) be an optimal solution of \eqref{primal}, and let \((\tilde v,\tilde r)\) be an optimal solution of its equality-constrained counterpart \eqref{primal_eq}. Assume that~(P$_{=}$) is feasible. Then $r^\star \ge \tilde r$.
Consequently, for every \(i=1,\dots,m\), we have $\hat g_i^\top v^\star \ge \hat g_i^\top \tilde v$.
\end{theorem}

Theorem~\ref{thm:config_comparison} shows that, for every objective, our method dominates ConFIG with respect to the rate of change along the computed direction. 
Moreover, the improvement can be strictly larger when some inequality constraints in \eqref{primal} are inactive at the optimum, as illustrated in the following example. 

\begin{example}
Let \(g_1 = c_1(1,0,0)\), \(g_2 = c_2(0,1,0)\), and \(g_3 = c_3(2,2,1)/3\), where \(c_1,c_2,c_3>0\). Then the normalized gradients are \(\hat g_1 = (1,0,0)\), \(\hat g_2 = (0,1,0)\), and \(\hat g_3 = (2,2,1)/3\). For \eqref{primal_eq}, the optimal direction is $\tilde v = (1,1,-1)/\sqrt{3}$, which gives rise to the ConFIG direction. In contrast, for \eqref{primal}, the optimal direction is $v^\star = (1,1,0)/\sqrt{2}$. Moreover, the corresponding inner products satisfy \(\hat g_1^\top \tilde v = \hat g_2^\top \tilde v = \hat g_3^\top \tilde v = 1/\sqrt{3}\), whereas \(\hat g_1^\top v^\star = \hat g_2^\top v^\star = 1/\sqrt{2}\) and \(\hat g_3^\top v^\star = 2\sqrt{2}/3\). Therefore, our method yields strictly larger rates of change for all three objectives in this example. This illustrates that enforcing equalities for all inner products, as in ConFIG, can be more restrictive than the inequality-based formulation underlying our method.
\end{example}

We also note that in the two-objective case, the optimum of \eqref{primal} necessarily activates both constraints, so our method and ConFIG coincide. Overall, ConFIG's property-driven construction imposes stronger conditions than necessary. In contrast, the Chebyshev-center geometry naturally recovers the intended properties in a relaxed form, where equality is required only for active constraints. 

\subsection{Comparison with minimum-norm methods based on the dual formulation}

We next compare our formulation with minimum-norm convex-combination methods. In particular, we compare with MGDA and GAPO, which construct update directions by minimizing the norm of a convex combination of gradients or normalized gradients.

\paragraph{MGDA}
MGDA~\cite{sener2018multi} determines the update direction as the minimum-norm convex combination of the original gradients. In our notation, its formulation can be written as $\min\{\|
\sum_{i=1}^m \alpha_i g_i
\|_2:
\sum_{i=1}^m \alpha_i = 1,
\alpha_i \ge 0, i=1,\dots,m.\}$.
This formulation is closely related to our dual formulation \eqref{dual}. The key difference is that MGDA uses the original gradients \(g_i\), whereas our dual uses the normalized gradients \(\hat g_i = g_i / \|g_i\|_2\). This distinction is particularly important for training PINNs. In PINNs, different loss terms often have substantially different scales, and such imbalance is known to interfere with training~\cite{wang2021understanding}. Because MGDA is based on the original gradients, loss-scale imbalance can directly affect the selected direction. 
In contrast, our dual formulation determines the direction using normalized gradients, thereby removing the influence of gradient magnitudes. Such magnitude-independent direction selection is well suited to PINN training.

\paragraph{GAPO}
GAPO~\cite{li2025gradient} normalizes the gradients in the MGDA objective to reduce influence of loss-scale imbalance.
In our notation, the formulation can be written as $\min\{\|
\sum_{i=1}^m \alpha_i g_i/\|g_i\|_2^\rho
\|_2:
\sum_{i=1}^m \alpha_i = 1,
\alpha_i \ge 0, i=1,\dots,m.\}$ 
where \(\rho\) controls the degree of gradient normalization. When \(\rho=1\), the normalized gradients become \(\hat g_i=g_i/\|g_i\|_2\), so GAPO coincides with the dual problem \eqref{dual} of our method. 
The difference lies in the derivation and the resulting hyperparameter dependence: GAPO introduces \(\rho\) as a tunable normalization exponent in the MGDA objective, whereas in our formulation the corresponding normalization is determined by the Chebyshev-center geometry and therefore does not depend on such a hyperparameter.

\section{Numerical experiments}
\label{sec:experiments}

We conduct numerical experiments to evaluate the proposed method in both forward and inverse PINN settings. We first test its performance on five forward PDE benchmark problems and compare it with a broad set of optimization baselines, including Adam~\cite{kingma2014adam}, LRA~\cite{wang2021understanding}, ReLoBRaLo~\cite{bischof2025multi}, MGDA~\cite{sener2018multi}, PCGrad~\cite{yu2020gradient}, IMTL-G~\cite{liu2021towards}, DCGD~\cite{hwang2024dual}, ConFIG~\cite{liu2025config}, and HARMONIC~\cite{bu2026harmonized}. 

We further consider an inverse problem focused on coefficient recovery under different data conditions, comparing the proposed method with the same set of baselines where applicable. We run each experiment over three independent trials with different random seeds and report the mean and standard deviation of the final relative \(L^2\) error. For forward problems, the error is computed on the solution, whereas for inverse problems, it is computed on the recovered coefficient.

\subsection{Comparison on forward benchmark equations}
\label{sec:forward_experiments}

We evaluate the \(\ell_2\)-norm formulation of the proposed method on five forward PDE benchmarks solved with vanilla PINNs: two steady-state boundary-value problems (Helmholtz and 2D Kovasznay flow) and three time-dependent initial-boundary-value problems (Burgers' equation, the 2D varying-coefficient heat equation, and the Klein--Gordon equation). We compare it against the baselines listed above; implementation details are provided in Appendix~\ref{app:forward_details}.
Table~\ref{tab:main_results} reports the mean and standard deviation of the final relative \(L^2\) errors. Lower values indicate better accuracy. In each column, we highlight the \textbf{best} and \underline{second-best} results. A dash denotes an omitted entry: DCGD is omitted for three-term objectives, and ConFIG is omitted for two-term objectives where it coincides with ours.

\begin{table}[t]
\centering
\caption{Final relative \(L^2\) errors on five forward PDE benchmark problems. Each value reports the mean and standard deviation over three independent runs with different random seeds.}
\label{tab:main_results}
\small
\begin{tabular}{lccccc}
\toprule
Method 
& Helmholtz 
& Kovasznay 
& Burgers' 
& Heat 2D VC 
& Klein--Gordon \\
\midrule
Adam      & 0.0307 (0.0162)  & 0.0044 (0.0010)  & 0.1659 (0.0358)  & 2.9813 (0.6427)  & 0.0267 (0.0111)  \\
LRA       & 1.0000 (0.0000)  & 0.0087 (0.0014)  & 0.4032 (0.0190)  & 0.9999 (0.0000)  & 1.0174 (0.0413)  \\
ReLoBRaLo & 0.0570 (0.0295)  & 0.0048 (0.0004)  & 0.1561 (0.0465)  & 2.6150 (0.4534)  & 0.0389 (0.0230)  \\
MGDA      & 0.7542 (0.4058)  & 0.0095 (0.0014)  & 0.1258 (0.0114)  & 0.9370 (0.0219)  & 0.1494 (0.0478)  \\
PCGrad    & \underline{0.0075 (0.0014)}  & 0.0057 (0.0012)  & 0.0706 (0.0046)  & 0.4185 (0.0829)  & 0.0466 (0.0194)  \\
IMTL-G    & 0.0086 (0.0007)  & \underline{0.0039 (0.0007)}  & 0.2591 (0.0235)  & 1.0099 (0.0128)  & 0.0871 (0.0950)  \\
DCGD      & 0.0097 (0.0061)  & 0.0052 (0.0025)  & \textemdash  & \textemdash  & \textemdash  \\
ConFIG    & \textemdash  & \textemdash  & 0.0708 (0.0191)  & 0.2401 (0.0359)  & 0.0339 (0.0068)  \\
HARMONIC  & 0.0116 (0.0060)  & 0.0061 (0.0045)  & \underline{0.0677 (0.0065)}  & \underline{0.2361 (0.0124)}  & \textbf{0.0143 (0.0031)}  \\
\midrule
Ours      & \textbf{0.0072 (0.0033)}  & \textbf{0.0017 (0.0005)}  & \textbf{0.0599 (0.0042)}  & \textbf{0.2316 (0.0190)}  & \underline{0.0218 (0.0074)}  \\
\bottomrule
\end{tabular}
\end{table}

Overall, these results demonstrate the superior empirical performance of the proposed method on both steady-state and time-dependent forward PDE problems.

\subsection{Coefficient recovery in an inverse heat problem}
\label{sec:inverse_experiments}

We further evaluate the proposed method on an inverse heat problem, where the goal is to recover the unknown diffusion coefficient \(a(x,y)\) from noisy observations of \(u(x,y,t)\). We consider three data settings: Standard, Higher noise, and Fewer data. We compare against the same baselines listed above except DCGD, with further implementation details provided in Appendix~\ref{app:inverse_details}.

Figure~\ref{fig:inverse_heat_reconstruction} compares the coefficient reconstructions obtained by Adam and our method. Adam produces a visibly distorted coefficient field with large structured errors, whereas our method recovers the ground-truth coefficient field with uniformly small reconstruction errors. Table~\ref{tab:inverse_heat_results} reports the mean and standard deviation of the final relative \(L^2\) errors for the recovered coefficient \(a\). Lower values indicate better coefficient recovery. In each column, we highlight the \textbf{best} and \underline{second-best} results. Overall, our method shows strong coefficient-recovery performance across the tested data settings.

\begin{figure}[t]
\centering
\includegraphics[width=\textwidth]{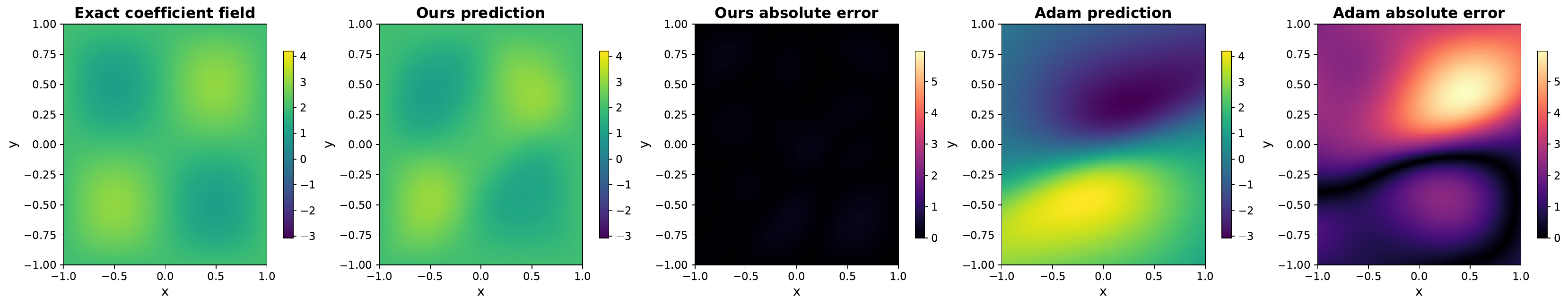}
\caption{Coefficient field reconstruction for the inverse heat problem. We visualize the ground-truth coefficient field, the reconstructed coefficient fields obtained by Adam and the proposed method, and their absolute reconstruction errors.}
\label{fig:inverse_heat_reconstruction}
\end{figure}

\begin{table}[t]
\centering
\caption{Final relative \(L^2\) errors of the recovered coefficient \(a\) for the inverse heat problem. Each value reports the mean and standard deviation over three independent runs
}
\label{tab:inverse_heat_results}
\small
\begin{tabular}{lccc}
\toprule
Method & Standard & Higher noise & Fewer data \\
\midrule
Adam      & 6.1510 (3.7224)  & 6.1409 (3.6977)  & 6.0555 (3.6322) \\
LRA       & 0.2390 (0.0000)  & 0.2390 (0.0000)  & 0.2390 (0.0000)  \\
ReLoBRaLo & 3.7730 (2.1329)  & 3.8501 (2.0031)  & 3.0611 (1.0022)  \\
MGDA      & 0.2389 (0.0001)  & 0.2389 (0.0000)  & 0.2389 (0.0000)  \\
PCGrad    & 0.1371 (0.0529)  & 0.1272 (0.0494)  & 0.1247 (0.0633) \\
IMTL-G    & 0.0599 (0.0005)  & 0.1069 (0.0085)  & 0.0686 (0.0026)  \\
ConFIG    & 0.0612 (0.0024)  & \underline{0.1038 (0.0042)}  & \underline{0.0655 (0.0034)}  \\
HARMONIC  & \underline{0.0596 (0.0014)}  & \textbf{0.0988 (0.0110)}  & 0.0694 (0.0008)  \\
\midrule
Ours      & \textbf{0.0593 (0.0002)}  & 0.1040 (0.0049)  & \textbf{0.0631 (0.0026)}  \\
\bottomrule
\end{tabular}
\end{table}

\section{Conclusions}
\label{sec:conclusions}

We proposed a Chebyshev center-based direction selection framework for multi-loss PINN training. We formulated update direction selection as finding a central direction that maximizes the minimum distance to the dual cone facets under a general $\ell_p/\ell_q$ geometry. We derived its low-dimensional dual problem, showed how to recover the corresponding primal update direction, and obtained an efficient algorithm with a convergence guarantee in the nonconvex setting. The geometric interpretation and primal--dual formulations also enabled a systematic comparison with representative direction-selection methods, clarifying how our approach relates to existing constructions. Experiments on forward and inverse PINN benchmarks demonstrated the effectiveness of the proposed method.

This work also leaves several directions for future study. Although the formulation is developed for general $\ell_p/\ell_q$ geometries, our experiments focused on the Euclidean case. A deeper theoretical and empirical understanding of how the choice of $p$ interacts with PDE structure, loss landscapes, and training dynamics would be valuable. Beyond PINNs, the same dual-cone and Chebyshev-center perspectives may also be useful for other multi-objective learning problems, including multi-task learning and recent applications such as multi-objective LLM alignment.

\bibliographystyle{unsrtnat}
\bibliography{ref}

\clearpage
\appendix

\section{Additional detail and proofs for section~\ref{sec:method}}
\label{app:method_proofs}

\subsection{Proof of proposition~\ref{prop:dual_formulation}}
\label{app:proof_dual_formulation}

Recall the primal problem
\begin{equation*}
\begin{aligned}
\max_{v,r}\quad & r\\
\text{s.t.}\quad & (\hat g_i^{(p)})^\top v \ge r,\quad i=1,\dots,m,\\
& \|v\|_q \le 1.
\end{aligned}
\tag{P}
\end{equation*}

We first derive its dual. Rewriting the constraints as
\[
(\hat g_i^{(p)})^\top v-r \ge 0,\quad i=1,\dots,m,
\qquad
1-\|v\|_q \ge 0,
\]
the Lagrangian is
\[
\mathcal L(v,r,\alpha,\beta)
=
r+\sum_{i=1}^m \alpha_i\left( (\hat g_i^{(p)})^\top v-r \right) +\beta(1-\|v\|_q),
\]
where \(\alpha_i\ge 0\) for all \(i\) and \(\beta\ge 0\). Rearranging terms gives
\[
\mathcal L(v,r,\alpha,\beta)
=
r\Bigl(1-\sum_{i=1}^m \alpha_i\Bigr)
+
\Bigl(\sum_{i=1}^m \alpha_i \hat g_i^{(p)}\Bigr)^\top v
-\beta \|v\|_q+\beta.
\]

Hence the dual function is
\[
\phi(\alpha,\beta)
=
\sup_{v,r}\mathcal L(v,r,\alpha,\beta).
\]

For the supremum over \(r\) to be finite, the coefficient of \(r\) must vanish. Therefore,
\[
\sum_{i=1}^m \alpha_i = 1.
\]
Otherwise, if \(1-\sum_i\alpha_i>0\), then letting \(r\to +\infty\) yields
\(\phi(\alpha,\beta)=+\infty\), and if \(1-\sum_i\alpha_i<0\), then letting
\(r\to -\infty\) yields \(\phi(\alpha,\beta)=+\infty\).

Next, we examine when the supremum over \(v\) is finite. Define
\[
w:=\sum_{i=1}^m\alpha_i\hat g_i^{(p)}.
\]
Then the part of the Lagrangian involving \(v\) is
\[
w^\top v-\beta\|v\|_q.
\]
By Hölder's inequality,
\[
w^\top v \le \|w\|_p\|v\|_q.
\]
Thus,
\[
w^\top v-\beta\|v\|_q
\le
(\|w\|_p-\beta)\|v\|_q.
\]    
Therefore, if \(\|w\|_p>\beta\), then there exists \(s\) with \(\|s\|_q=1\) and
\(w^\top s>\beta\). Hence, along \(v=\tau s\),
\[
w^\top v-\beta\|v\|_q
=
\tau(w^\top s-\beta)\to+\infty,
\]
so the supremum over \(v\) is \(+\infty\).
Thus, for the dual function to be finite, it is necessary that
\[
\|w\|_p\le \beta.
\]

Under this condition, we now compute the value of the supremum. If \(\|w\|_p\le \beta\), then
\[
w^\top v-\beta\|v\|_q\le 0
\]
for all \(v\), and equality is attained at \(v=0\). Hence
\[
\sup_v \bigl(w^\top v-\beta\|v\|_q\bigr)=0
\qquad\text{whenever } \|w\|_p\le \beta.
\]

Therefore, the dual function is
\[
\phi(\alpha,\beta)=\beta
\]
subject to
\[
\sum_{i=1}^m \alpha_i=1,\qquad
\Bigl\|\sum_{i=1}^m \alpha_i\hat g_i^{(p)}\Bigr\|_p\le \beta,\qquad
\alpha_i\ge 0,\ \beta\ge 0.
\]
Minimizing over \(\beta\) yields
\[
\beta^\star=
\Bigl\|\sum_{i=1}^m \alpha_i\hat g_i^{(p)}\Bigr\|_p,
\]
and thus the dual problem reduces to
\begin{equation*}
\begin{aligned}
\min_{\alpha}\quad &
\Bigl\|\sum_{i=1}^m \alpha_i\hat g_i^{(p)}\Bigr\|_p\\
\text{s.t.}\quad &
\sum_{i=1}^m \alpha_i=1,\\
& \alpha_i\ge 0,\quad i=1,\dots,m.
\end{aligned}
\tag{D}
\end{equation*}

It remains to show strong duality. The primal \eqref{primal} can be written as the minimization of \(-r\) subject to
\[
r-(\hat g_i^{(p)})^\top v\le 0,\qquad i=1,\dots,m,
\]
and
\[
\|v\|_q \le 1.
\] 
The objective \(-r\) is linear, each inequality 
\(r-(\hat g_i^{(p)})^\top v\le 0\) is affine in \((v,r)\), and the constraint \(\|v\|_q\le 1\) defines a convex set. Hence, \eqref{primal} is a convex optimization problem in the equivalent minimization form. Moreover, Slater's condition holds, because \((v,r)=(0,-1)\) is strictly feasible:
\[
r-(\hat g_i^{(p)})^\top v=-1<0,\quad i=1,\dots,m,
\qquad
\|v\|_q = 0 <1.
\]
Therefore, by Slater's theorem, strong duality holds between~(P) and~(D).
\qed

\subsection{Proof of proposition~\ref{prop:recovery_direction}}
\label{app:proof_recovery_direction}

Let \((v^\star,r^\star)\) be an optimal solution of the primal \eqref{primal}, and let
\(\alpha^\star\) be an optimal solution of the dual \eqref{dual}. Define
\[
w^\star := \sum_{i=1}^m \alpha_i^\star \hat g_i^{(p)},
\]
and let \(\beta^\star := \|w^\star\|_p\).

Assume that \(r^\star>0\). Then \(v^\star\neq 0\), since if \(v^\star=0\), feasibility of \((v^\star,r^\star)\) would imply
\[
(\hat g_i^{(p)})^\top v^\star = 0 \ge r^\star
\]
for all \(i\), and hence \(r^\star\le 0\), a contradiction.

From the proof of Proposition~\ref{prop:dual_formulation}, the Lagrangian is
\[
\mathcal L(v,r,\alpha,\beta)
=
r+\sum_{i=1}^m \alpha_i\bigl((\hat g_i^{(p)})^\top v-r\bigr)+\beta(1-\|v\|_q).
\]
Since \(1<q<\infty\), the \(\ell_q\) norm is differentiable at \(v^\star\neq 0\). Therefore, the KKT stationarity condition with respect to \(v\) gives
\[
\sum_{i=1}^m \alpha_i^\star \hat g_i^{(p)} - \beta^\star \nabla_v \|v^\star\|_q=0.
\]
Equivalently,
\[
w^\star = \beta^\star \nabla_v \|v^\star\|_q.
\tag{1}
\]

We next show that \(\|v^\star\|_q=1\). By strong duality, the primal and dual optimal values coincide.
From Proposition~\ref{prop:dual_formulation}, the dual optimal value is \(\beta^\star\), while the primal
optimal value is \(r^\star\). Hence
\[
\beta^\star = r^\star.
\]
Under the assumption \(r^\star>0\), we therefore have \(\beta^\star>0\) and \(w^\star\neq0\).

Now consider the complementary slackness condition associated with the constraint
\(\|v\|_q\le 1\):
\[
\beta^\star(1-\|v^\star\|_q)=0.
\]
Since \(\beta^\star>0\), it follows that
\[
\|v^\star\|_q=1.
\tag{2}
\]

Since the gradient of the \(\ell_q\) norm at \(v^\star\) is
\[
\nabla_v \|v^\star\|_q
=
\frac{\operatorname{sgn}(v^\star)\odot |v^\star|^{q-1}}
{\|v^\star\|_q^{q-1}},
\]
where the operations are applied componentwise, and \(\|v^\star\|_q=1\) by \((2)\), equation~\((1)\) becomes
\[
w^\star
=
\beta^\star
\bigl(\operatorname{sgn}(v^\star)\odot |v^\star|^{q-1}\bigr).
\tag{3}
\]

Therefore, componentwise,
\[
v_j^\star
=
\operatorname{sgn}(w_j^\star)
\left(\frac{|w_j^\star|}{\beta^\star}\right)^{1/(q-1)}.
\]
Since \(q=p/(p-1)\), we have \(1/(q-1)=p-1\), and hence
\[
v_j^\star
=
\frac{
\operatorname{sgn}(w_j^\star) |w_j^\star|^{p-1}
}{
(\beta^\star)^{p-1}
}.
\]

Since \(\beta^\star=\|w^\star\|_p\), this gives
\[
v^\star
=
\frac{
\operatorname{sgn}(w^\star)\odot |w^\star|^{p-1}
}{
\|w^\star\|_p^{p-1}
}.
\]

This completes the proof.
\qed

\subsection{The \(p=1\) and \(p=\infty\) cases for the recovery step}
\label{app:p1_pinf_recovery}

The main text focuses on \(1<p<\infty\), where the primal direction is uniquely recovered from the dual solution in closed form, as shown in Proposition~\ref{prop:recovery_direction}. The cases \(p=1\) and \(p=\infty\) are also well defined for both the primal and the dual problem. The difference is that the recovery step is no longer given by a closed-form map from the dual solution to a unique primal direction. Indeed, when \(p=1\) or \(p=\infty\), the support direction associated with a dual vector can be non-unique, because the corresponding unit $\ell_q$ ball has flat faces.

Given an optimal dual solution \(\alpha^\star\), define
\[
w^\star := \sum_{i=1}^m \alpha_i^\star \hat g_i^{(p)},
\qquad
r^\star := \|w^\star\|_p .
\]
By strong duality, \(r^\star\) is the optimal value of the primal problem. Hence, a valid primal optimal direction can be recovered by finding \(v^\star\) such that
\[
(\hat g_i^{(p)})^\top v^\star \ge r^\star,
\qquad i=1,\dots,m,
\]
and
\[
\|v^\star\|_q \le 1.
\]
Any \(v^\star\) satisfying these conditions is primal optimal, since it attains the optimal value \(r^\star\). Thus, in the \(p=1\) and \(p=\infty\) cases, the dual solution provides the optimal radius \(r^\star\), while the corresponding primal direction can be selected from the set of directions that attain this radius.

For \(p=1\), we have \(q=\infty\). In this case, the norm constraint becomes
\[
\|v\|_\infty \le 1
\quad\Longleftrightarrow\quad
-1 \le v_j \le 1,\qquad j=1,\dots,n.
\]
Therefore, the recovery step can be written as the linear feasibility problem
\[
(\hat g_i^{(1)})^\top v \ge r^\star,
\qquad i=1,\dots,m,
\]
\[
-1 \le v_j \le 1,
\qquad j=1,\dots,n.
\]
Equivalently, \(v^\star\) can be obtained directly by solving the primal problem \eqref{primal} as the following linear program:
\[
\begin{aligned}
\max_{v,r}\quad & r \\
\text{s.t.}\quad
& (\hat g_i^{(1)})^\top v \ge r,
\quad i=1,\dots,m,\\
& -1 \le v_j \le 1,
\quad j=1,\dots,n .
\end{aligned}
\]

For \(p=\infty\), we have \(q=1\). The recovery step is
\[
(\hat g_i^{(\infty)})^\top v \ge r^\star,
\qquad i=1,\dots,m,
\qquad
\|v\|_1 \le 1.
\]
Introducing auxiliary variables \(z_j\ge 0\), the \(\ell_1\)-constraint can be written as
\[
-z_j \le v_j \le z_j,
\qquad j=1,\dots,n,
\]
\[
\sum_{j=1}^n z_j \le 1.
\]
Thus, the recovery step becomes the linear feasibility problem
\[
(\hat g_i^{(\infty)})^\top v \ge r^\star,
\qquad i=1,\dots,m,
\]
\[
-z_j \le v_j \le z_j,
\qquad j=1,\dots,n,
\]
\[
\sum_{j=1}^n z_j \le 1,
\qquad
z_j\ge 0,\quad j=1,\dots,n.
\]
Similarly, \(v^\star\) can be obtained directly by solving the primal problem \eqref{primal} as the following linear program:
\[
\begin{aligned}
\max_{v,r,z}\quad & r \\
\text{s.t.}\quad
& (\hat g_i^{(\infty)})^\top v \ge r,
\quad i=1,\dots,m,\\
& -z_j \le v_j \le z_j,
\quad j=1,\dots,n,\\
& \sum_{j=1}^n z_j \le 1,\\
& z_j\ge 0,
\quad j=1,\dots,n .
\end{aligned}
\]

These formulations show that the \(p=1\) and \(p=\infty\) cases can be handled through LP-based direction selection. However, unlike the \(1<p<\infty\) case, the primal direction is not recovered from \(w^\star\) by a single closed-form map, and the LP formulation may still have multiple optimal directions. When this happens, the Chebyshev-center criterion alone does not select a unique direction, so an additional selection criterion may be required.

\subsection{Proof of proposition~\ref{prop:r0_stationary}}
\label{app:proof_r0_stationary}

Let \((v^\star,r^\star)\) be an optimal solution of the primal \eqref{primal} at \(\theta\), and assume that
\[
r^\star=0.
\]
Let \(\alpha^\star\) be an optimal solution of the dual \eqref{dual}. By strong duality from
Proposition~\ref{prop:dual_formulation}, the primal and dual optimal values coincide. Hence,
\[
\left\|\sum_{i=1}^m \alpha_i^\star \hat g_i^{(p)}\right\|_p = 0,
\]
which implies
\[
\sum_{i=1}^m \alpha_i^\star \hat g_i^{(p)} = 0.
\]
Since \(\hat g_i^{(p)} = g_i/\|g_i\|_p\), where \(g_i=\nabla \mathcal{L}_i(\theta)\), we obtain
\[
\sum_{i=1}^m \frac{\alpha_i^\star}{\|g_i\|_p} \nabla \mathcal{L}_i(\theta) = 0.
\]

Now define
\[
\bar\alpha_i
:=
\frac{\alpha_i^\star/\|g_i\|_p}{\sum_{j=1}^m \alpha_j^\star/\|g_j\|_p},
\qquad i=1,\dots,m.
\]
Since \(\alpha_i^\star\ge 0\) and \(\sum_{i=1}^m \alpha_i^\star=1\), we have \(\bar\alpha_i\ge 0\) for all \(i\) and
\[
\sum_{i=1}^m \bar\alpha_i = 1.
\]
Moreover,
\[
\sum_{i=1}^m \bar\alpha_i \nabla \mathcal{L}_i(\theta)
=
\frac{1}{\sum_{j=1}^m \alpha_j^\star/\|g_j\|_p}
\sum_{i=1}^m \frac{\alpha_i^\star}{\|g_i\|_p} \nabla \mathcal{L}_i(\theta)
=0.
\]
Therefore, there exists \(\bar\alpha\in\mathbb{R}_+^m\) such that
\[
\sum_{i=1}^m \bar\alpha_i=1
\qquad\text{and}\qquad
\sum_{i=1}^m \bar\alpha_i \nabla \mathcal{L}_i(\theta)=0.
\]
By Definition~\ref{def:pareto_stat}, \(\theta\) is Pareto-stationary.
\qed

\subsection{Solving the dual problems for \(p=1\) and \(p=\infty\)}
\label{app:p1_pinf_dual}

The dual problem \eqref{dual} remains convex at the cases \(p=1\) and \(p=\infty\), but the objective is nonsmooth. Instead of applying the
Frank--Wolfe step in Algorithm~\ref{alg:dual-fw} directly, these 
dual problems can be written as linear programs.

For \(p=1\), the dual problem is
\[
\min_{\alpha\in\Delta^m}
\left\|
\sum_{i=1}^m \alpha_i \hat g_i^{(1)}
\right\|_1 .
\]
Using auxiliary variables \(z_j\ge 0\), this is equivalent to
\[
\begin{aligned}
\min_{\alpha,z}\quad & \sum_{j=1}^n z_j \\
\text{s.t.}\quad
& -z_j
\le
\left(\sum_{i=1}^m \alpha_i \hat g_i^{(1)}\right)_j
\le
z_j,
\quad j=1,\dots,n,\\
& \sum_{i=1}^m \alpha_i=1,\\
& \alpha_i\ge0,\quad i=1,\dots,m,\\
& z_j\ge0,\quad j=1,\dots,n.
\end{aligned}
\]
The optimal value is then
\[
r^\star=\sum_{j=1}^n z_j^\star.
\]

For \(p=\infty\), the dual problem is
\[
\min_{\alpha\in\Delta^m}
\left\|
\sum_{i=1}^m \alpha_i \hat g_i^{(\infty)}
\right\|_\infty .
\]
Introducing a scalar variable \(\tau\), this is equivalent to
\[
\begin{aligned}
\min_{\alpha,\tau}\quad & \tau \\
\text{s.t.}\quad
& -\tau
\le
\left(\sum_{i=1}^m \alpha_i \hat g_i^{(\infty)}\right)_j
\le
\tau,
\quad j=1,\dots,n,\\
& \sum_{i=1}^m \alpha_i=1,\\
& \alpha_i\ge0,\quad i=1,\dots,m,\\
& \tau\ge0.
\end{aligned}
\]
The optimal value is
\[
r^\star=\tau^\star.
\]

Thus, in both \(p=1\) and \(p=\infty\) cases, an optimal dual solution \(\alpha^\star\)
and the corresponding optimal value \(r^\star\) can be obtained by solving
a linear program.

\subsection{Proof of theorem~\ref{thm:convergence}}
\label{app:proof_convergence}

For each iteration \(t\), let \(\alpha_t\) be an optimal solution of the dual \eqref{dual} at \(\theta_t\), and define
\[
w_t:=\sum_{i=1}^m \alpha_{t,i}\hat g_i^{(p)}(\theta_t),
\qquad
v_t:=\frac{\operatorname{sgn}(w_t)\odot |w_t|^{p-1}}{\|w_t\|_p^{p-1}},
\qquad
d_t:=\Bigl(\sum_{i=1}^m g_i(\theta_t)^\top v_t\Bigr)v_t.
\]
Also let
\[
\nabla \mathcal{L}(\theta_t)=\sum_{i=1}^m g_i(\theta_t),
\]
and let \(r_t^\star\) denote the optimal value of the primal \eqref{primal} at \(\theta_t\).

If Algorithm~\ref{alg:dual-fw} terminates at some \(t\in\{0,\dots,T-1\}\), then it returns an
\(\epsilon\)-Pareto-stationary point by construction. We therefore consider the remaining case where the algorithm does not terminate for any \(t=0,\dots,T-1\). Then
\[
\|w_t\|_p>\epsilon
\qquad\text{for all } t=0,\dots,T-1.
\]
By strong duality, \(\|w_t\|_p=r_t^\star\). Hence
\[
r_t^\star>\epsilon
\qquad\text{for all } t=0,\dots,T-1.
\tag{1}
\]

We first establish two auxiliary relations used below.

\noindent\textbf{Step 1: Lower bound on \(\|d_t\|_q\).}

Since \(\alpha_t\) is optimal for the dual \eqref{dual} and \(r_t^\star>0\) by~(1), Proposition~\ref{prop:recovery_direction} implies that \(v_t\) is the corresponding optimal primal direction for~(P). Therefore,
\[
\bigl(\hat g_i^{(p)}(\theta_t)\bigr)^\top v_t \ge r_t^\star,
\qquad i=1,\dots,m.
\]
Multiplying both sides by \(\|g_i(\theta_t)\|_p\) yields
\[
g_i(\theta_t)^\top v_t \ge r_t^\star \|g_i(\theta_t)\|_p,
\qquad i=1,\dots,m.
\]
Summing over \(i\), we obtain
\[
\sum_{i=1}^m g_i(\theta_t)^\top v_t
\ge
\sum_{i=1}^m r_t^\star \|g_i(\theta_t)\|_p
= r_t^\star\sum_{i=1}^m \|g_i(\theta_t)\|_p
\ge r_t^\star \Bigl\|\sum_{i=1}^m g_i(\theta_t)\Bigr\|_p
= r_t^\star \|\nabla \mathcal{L}(\theta_t)\|_p,
\]
where the second inequality follows from the triangle inequality. Since \(\|v_t\|_q=1\) and \(\sum_{i=1}^m g_i(\theta_t)^\top v_t \ge 0\), it follows that
\[
\|d_t\|_q
=
\|\Bigl(\sum_{i=1}^m g_i(\theta_t)^\top v_t\Bigr)v_t\|_q
=
\sum_{i=1}^m g_i(\theta_t)^\top v_t
\ge
r_t^\star \|\nabla \mathcal{L}(\theta_t)\|_p.
\]
Combining this with \((1)\) gives
\[
\|d_t\|_q^2 \ge \epsilon^2 \|\nabla \mathcal{L}(\theta_t)\|_p^2.
\tag{2}
\]

\noindent\textbf{Step 2: Relation between \(\nabla \mathcal{L}(\theta_t)\) and \(d_t\).}

By definition,
\[
\nabla \mathcal{L}(\theta_t)^\top d_t
=
\Bigl(\sum_{i=1}^m g_i(\theta_t)\Bigr)^\top \Bigl(\sum_{i=1}^m g_i(\theta_t)^\top v_t\Bigr)v_t
=
\Bigl(\sum_{i=1}^m g_i(\theta_t)^\top v_t\Bigr)^2.
\]
On the other hand, since \(\|v_t\|_q=1\),
\[
\|d_t\|_q^2
= \|\Bigl(\sum_{i=1}^m g_i(\theta_t)^\top v_t\Bigr)v_t\|_q^2
= \Bigl(\sum_{i=1}^m g_i(\theta_t)^\top v_t\Bigr)^2.
\]
Therefore,
\[
\nabla \mathcal{L}(\theta_t)^\top d_t=\|d_t\|_q^2.
\tag{3}
\]

\noindent\textbf{Step 3: Descent estimate and final bound via telescoping.}

Now use the \(\beta\)-smoothness of \(\mathcal{L}\) with respect to \(\|\cdot\|_q\). Since
\[
\theta_{t+1}=\theta_t-\eta d_t,
\]
we have
\[
\mathcal{L}(\theta_{t+1})
\le
\mathcal{L}(\theta_t)
-\eta \nabla \mathcal{L}(\theta_t)^\top d_t
+\frac{\beta\eta^2}{2}\|d_t\|_q^2.
\]
Applying \((3)\), this becomes
\[
\mathcal{L}(\theta_{t+1})
\le
\mathcal{L}(\theta_t)
-\eta\Bigl(1-\frac{\beta\eta}{2}\Bigr)\|d_t\|_q^2.
\]
Now choose \(\eta=1/\beta\). Then
\[
\mathcal{L}(\theta_{t+1})
\le
\mathcal{L}(\theta_t)-\frac{1}{2\beta}\|d_t\|_q^2.
\]
Using \((2)\), we further obtain
\[
\mathcal{L}(\theta_{t+1})
\le
\mathcal{L}(\theta_t)-\frac{\epsilon^2}{2\beta}\|\nabla \mathcal{L}(\theta_t)\|_p^2.
\tag{4}
\]

Summing \((4)\) over \(t=0,\dots,T-1\) yields
\[
\mathcal{L}(\theta_T)-\mathcal{L}(\theta_0)
\le
-\frac{\epsilon^2}{2\beta}\sum_{t=0}^{T-1} \|\nabla \mathcal{L}(\theta_t)\|_p^2.
\]
Rearranging,
\[
\sum_{t=0}^{T-1} \|\nabla \mathcal{L}(\theta_t)\|_p^2
\le
\frac{2\beta\bigl(\mathcal{L}(\theta_0)-\mathcal{L}(\theta_T)\bigr)}{\epsilon^2}.
\]
Since \(\mathcal{L}(\theta^\star)\le \mathcal{L}(\theta_t)\) for all \(t\), in particular
\(\mathcal{L}(\theta^\star)\le \mathcal{L}(\theta_T)\). Hence
\[
\sum_{t=0}^{T-1} \|\nabla \mathcal{L}(\theta_t)\|_p^2
\le
\frac{2\beta\bigl(\mathcal{L}(\theta_0)-\mathcal{L}(\theta^\star)\bigr)}{\epsilon^2}.
\]
Dividing by \(T\) and using
\[
\min_{0\le t\le T-1}\|\nabla \mathcal{L}(\theta_t)\|_p^2
\le
\frac{1}{T}\sum_{t=0}^{T-1} \|\nabla \mathcal{L}(\theta_t)\|_p^2,
\]
we conclude that
\[
\min_{0\le t\le T-1}\|\nabla \mathcal{L}(\theta_t)\|_p^2
\le
\frac{2\beta\bigl(\mathcal{L}(\theta_0)-\mathcal{L}(\theta^\star)\bigr)}{\epsilon^2T}.
\]

Therefore, unless Algorithm~\ref{alg:dual-fw} terminates earlier at an \(\epsilon\)-Pareto-stationary point, the
best iterate among \(\{\theta_t\}_{t=0}^{T-1}\) satisfies the stated bound.
\qed

\section{Additional detail and proof for section~\ref{sec:compare}}
\label{app:compare_details}

\subsection{Details of the comparative toy example}
\label{app:comparison_toy}

We consider a three-objective toy example in \(\mathbb{R}^3\). Let
\[
g_1=(5,0,0),\qquad 
g_2=(0,3,0),\qquad
g_3=\left(\frac15,\frac{14}{15},\frac{2\sqrt5}{15}\right).
\]
Then \(\|g_1\|_2=5\), \(\|g_2\|_2=3\), and \(\|g_3\|_2=1\). The normalized gradients are
\[
\hat g_1=(1,0,0),\qquad
\hat g_2=(0,1,0),\qquad
\hat g_3=\left(\frac15,\frac{14}{15},\frac{2\sqrt5}{15}\right).
\]

MGDA solves
\[
\min_{\alpha\in\Delta_3}
\left\|
\sum_{i=1}^3 \alpha_i g_i
\right\|_2^2.
\]
For the gradients in this example, the solution is
\[
\alpha_{\rm MGDA}=(0,0,1),
\]
and hence
\[
v_{\rm MGDA}
=
g_3
=
\left(\frac15,\frac{14}{15},\frac{2\sqrt5}{15}\right).
\]

Although \(v_{\rm MGDA}\in \mathbf{K}^*\), it lies close to the facet of the dual cone. This shows that MGDA is affected by gradient-scale imbalance, causing the selected direction to be biased toward \(g_3\).

For IMTL-G, using its closed-form coefficient rule yields
\[
\alpha_{\rm IMTL\text{-}G}
=
\left(
-\frac{13}{22},
-\frac{20}{11},
\frac{75}{22}
\right).
\]
The resulting aggregated direction before normalization is
\[
\sum_{i=1}^3 \alpha_i g_i
=
\left(
-\frac{25}{11},
-\frac{25}{11},
\frac{5\sqrt5}{11}
\right).
\]
After normalization, this gives
\[
v_{\rm IMTL\text{-}G}
=
\left(
-\sqrt{\frac5{11}},
-\sqrt{\frac5{11}},
\frac1{\sqrt{11}}
\right).
\]
Therefore, the normalized inner products are
\[
\hat g_1^\top v_{\rm IMTL\text{-}G}
=
\hat g_2^\top v_{\rm IMTL\text{-}G}
=
\hat g_3^\top v_{\rm IMTL\text{-}G}
=
-\sqrt{\frac5{11}}.
\]
Thus \(v_{\rm IMTL\text{-}G}\notin \mathbf{K}^*\).

For ConFIG, the direction enforces equal normalized inner products,
\[
\hat g_1^\top v
=
\hat g_2^\top v
=
\hat g_3^\top v.
\]
Solving these equalities and normalizing the direction gives
\[
v_{\rm ConFIG}
=
\left(
\sqrt{\frac5{11}},
\sqrt{\frac5{11}},
-\frac1{\sqrt{11}}
\right).
\]
Indeed,
\[
\hat g_1^\top v_{\rm ConFIG}
=
\hat g_2^\top v_{\rm ConFIG}
=
\hat g_3^\top v_{\rm ConFIG}
=
\sqrt{\frac5{11}}.
\]
Therefore,
\[
\tilde r
=
\sqrt{\frac5{11}}
\approx 0.674.
\]

Our method solves
\[
\max_{\|v\|_2\le 1}\min_{i=1,2,3}\hat g_i^\top v.
\]
In this example, the solution is
\[
v_{\rm ours}
=
\left(\frac1{\sqrt2},\frac1{\sqrt2},0\right).
\]
The normalized inner products are
\[
\hat g_1^\top v_{\rm ours}
=
\hat g_2^\top v_{\rm ours}
=
\frac1{\sqrt2},
\qquad
\hat g_3^\top v_{\rm ours}
=
\frac{17}{15\sqrt2}.
\]
Hence
\[
r^\star=\frac1{\sqrt2}\approx 0.707.
\]
The third constraint is inactive because
\[
\hat g_3^\top v_{\rm ours}
=
\frac{17}{15\sqrt2}
>
\frac1{\sqrt2}.
\]
Therefore,
\[
\tilde r
=
\sqrt{\frac5{11}}
<
\frac1{\sqrt2}
=
r^\star.
\]
This illustrates that enforcing equality for all objectives can be more restrictive than the max--min formulation.

\subsection{Derivation of the equivalence between ConFIG and the equality-constrained problem}
\label{app:config_equiv}

Suppose \(r>0\), and define a new variable \(w = v/r\). Then the equality constraints become
\[
\hat g_i^\top w = 1,\qquad i=1,\dots,m,
\]
or equivalently,
\[
\hat G^\top w = \mathbf{1}_m,
\qquad
\hat G = [\hat g_1,\hat g_2,\ldots,\hat g_m].
\]
Moreover, since \(v = rw\) and \(\|v\|_2 \le 1\), we have
\[
r\|w\|_2 = \|v\|_2 \le 1,
\]
which implies
\[
r \le \frac{1}{\|w\|_2}.
\]
Therefore, maximizing \(r\) in~(P$_{=}$) is equivalent to minimizing \(\|w\|_2\) subject to \(\hat G^\top w = \mathbf{1}_m\), that is,
\[
\min_w \|w\|_2
\qquad
\text{s.t.}\quad
\hat G^\top w = \mathbf{1}_m.
\]
The minimum-norm solution of this system is given by
\[
w^\star = (\hat G^\top)^+ \mathbf{1}_m.
\]
Hence, the optimal direction of~(P$_{=}$) is obtained by normalizing \(w^\star\):
\[
v^\star = \frac{w^\star}{\|w^\star\|_2}.
\]
Therefore, the normalized direction used by ConFIG coincides with the optimal direction of the equality-constrained counterpart of our primal \eqref{primal}. 

\subsection{Proof of theorem~\ref{thm:config_comparison}}
\label{app:proof_config_comparison}

Let \(\mathcal{F}\) and \(\mathcal{F}_{=}\) denote the feasible sets of Problems~(P) and~(P$_{=}$), respectively.

Since every equality implies the corresponding inequality, we have
\[
\mathcal{F}_{=} \subseteq \mathcal{F}.
\]
Therefore, \eqref{primal} is a relaxation of Problem~(P$_{=}$). Since both problems maximize the same objective \(r\), the optimal value over the larger feasible set cannot be smaller than that over the smaller one. Hence,
\[
r^\star \ge \tilde r.
\]

By feasibility of \((v^\star,r^\star)\) for~(P), we have
\[
\hat g_i^\top v^\star \ge r^\star,
\qquad i=1,\dots,m.
\]
Combining this with \(r^\star \ge \tilde r\), we obtain
\[
\hat g_i^\top v^\star \ge \tilde r,
\qquad i=1,\dots,m.
\]
Since \((\tilde v,\tilde r)\) is feasible for Problem~(P$_{=}$), its equality constraints imply
\[
\hat g_i^\top \tilde v = \tilde r,
\qquad i=1,\dots,m.
\]
Therefore,
\[
\hat g_i^\top v^\star \ge \hat g_i^\top \tilde v,
\qquad i=1,\dots,m.
\]

This completes the proof.
\qed

\section{Pseudocode for Algorithms}
\label{app:algorithms}

\begin{algorithm}[t]
\caption{Exact Dual Solver for Two Loss Terms}
\label{alg:exact-dual-two-loss}
\begin{algorithmic}[1]
\Require $\ell_2$-normalized gradients $\hat g_1,\hat g_2\in\mathbb{R}^n$
\Ensure Dual weights $\alpha^\star\in\Delta^2$

\State Set $\alpha^\star \gets (1/2,\,1/2)$
\State \Return $\alpha^\star$
\end{algorithmic}
\end{algorithm}

Algorithm~\ref{alg:exact-dual-two-loss} follows from the closed-form structure of
the two-loss dual problem. In this case, the dual objective reduces to minimizing
\[
\left\|\alpha \hat g_1 + (1-\alpha)\hat g_2\right\|_2^2
\quad\text{over}\quad \alpha\in[0,1].
\]
Since both gradients are $\ell_2$-normalized, this objective is minimized at
$\alpha=1/2$ whenever $\hat g_1\ne \hat g_2$. If $\hat g_1=\hat g_2$, every
$\alpha\in[0,1]$ gives the same direction, so the symmetric choice
$\alpha=1/2$ remains valid.

\begin{algorithm}[t]
\caption{Exact Dual Solver for Three Loss Terms}
\label{alg:exact-dual-three-loss}
\begin{algorithmic}[1]
\Require $\ell_2$-normalized gradients $\hat g_1,\hat g_2,\hat g_3\in\mathbb{R}^n$, feasibility tolerance $\tau>0$, numerical tolerance $\varepsilon>0$
\Ensure Dual weights $\alpha^\star\in\Delta^3$

\State Form the Gram matrix $G\in\mathbb{R}^{3\times 3}$ with $G_{ij}=\hat g_i^\top \hat g_j$

\State Find $\alpha^0\in\Delta^3$ satisfying $G\alpha^0=0$, if it exists
\If{such $\alpha^0$ exists}
    \State \Return $\alpha^0$
\EndIf

\State $z\gets G^\dagger \mathbf{1}$
\State $c\gets \mathbf{1}^\top z$
\If{$c>\varepsilon$}
    \State $\tilde\alpha\gets z/c$
    \If{$\tilde\alpha_i\ge -\tau$ for all $i=1,2,3$}
        \State $\tilde\alpha\gets (\tilde\alpha)_+$
        \State $\tilde\alpha\gets \tilde\alpha/(\mathbf{1}^\top\tilde\alpha)$
        \State \Return $\tilde\alpha$
    \EndIf
\EndIf

\State Find
\[
(i^\star,j^\star)\in
\arg\min_{1\le i<j\le 3}G_{ij}.
\]
\State $\alpha^\star\gets (0,0,0)$
\State $\alpha^\star_{i^\star}\gets 1/2$
\State $\alpha^\star_{j^\star}\gets 1/2$
\State \Return $\alpha^\star$
\end{algorithmic}
\end{algorithm}

Algorithm~\ref{alg:exact-dual-three-loss} uses the simple geometric structure of
the $\ell_2$ dual problem over $\Delta^3$. Its solution is either a zero-norm
convex combination, an interior simplex point, or an edge solution. The first
case corresponds to
$0\in\operatorname{conv}\{\hat g_1,\hat g_2,\hat g_3\}$. In the second case, the
KKT condition implies equal inner products with all three normalized gradients,
which yields the candidate
$\tilde\alpha=G^\dagger\mathbf{1}/(\mathbf{1}^\top G^\dagger\mathbf{1})$.
If this candidate is infeasible, the minimizer lies on an edge. Because the
gradients are $\ell_2$-normalized, the exact edge solution is the midpoint of the
pair with the smallest Gram entry $G_{ij}$.

\begin{algorithm}[t]
\caption{\textsc{DualChebyshev-Exact-Adam}}
\label{alg:dualchebyshev-exact-adam}
\begin{algorithmic}[1]
\Require Initial point $\theta_0$, step sizes $\{\eta_t\}_{t=0}^{T-1}$, Adam parameters $\beta_1,\beta_2,\epsilon_{\rm Adam}$, threshold $\delta>0$
\Ensure Updated parameters $\theta$

\State Initialize Adam moments $M_0\gets 0$, $S_0\gets 0$
\For{$t=0,1,\dots,T-1$}
    \State Compute grouped gradients $g_1(\theta_t),\dots,g_m(\theta_t)$
    \State Compute $\ell_2$-normalized gradients $\hat g_1(\theta_t),\dots,\hat g_m(\theta_t)$

    \If{$m=2$}
    \State Apply the exact two-loss solver in Algorithm~\ref{alg:exact-dual-two-loss} and obtain $\alpha_t\in\Delta^2$
\ElsIf{$m=3$}
    \State Apply the exact three-loss solver in Algorithm~\ref{alg:exact-dual-three-loss} and obtain $\alpha_t\in\Delta^3$
\Else
    \State Apply Frank--Wolfe to the dual \eqref{dual} and obtain $\alpha_t\in\Delta^m$
\EndIf

    \State Form
    \[
    w_t\gets \sum_{i=1}^m \alpha_{t,i}\hat g_i(\theta_t)
    \]
    \If{$\|w_t\|_2\le \delta$}
        \State \textbf{terminate} and return $\theta_t$
    \EndIf

    \State Recover the primal direction:
    \[
    v_t\gets \frac{w_t}{\|w_t\|_2}
    \]
    \State Set the final update direction with adaptive scalar:
    \[
    d_t\gets
    \left(\sum_{i=1}^m g_i(\theta_t)^\top v_t\right)v_t
    \]

    \State $M_{t+1}\gets \beta_1M_t+(1-\beta_1)d_t$
    \State $S_{t+1}\gets \beta_2S_t+(1-\beta_2)(d_t\odot d_t)$
    \State $\widehat M_{t+1}\gets M_{t+1}/(1-\beta_1^{t+1})$
    \State $\widehat S_{t+1}\gets S_{t+1}/(1-\beta_2^{t+1})$
    \State $\theta_{t+1}\gets \theta_t-\eta_t\,\widehat M_{t+1}/(\sqrt{\widehat S_{t+1}}+\epsilon_{\rm Adam})$
\EndFor
\State \Return $\theta_T$
\end{algorithmic}
\end{algorithm}

Algorithm~\ref{alg:dualchebyshev-exact-adam} combines the exact dual solvers above
with an Adam-style parameter update. When $m=2$ or $m=3$, the dual weights are
computed by Algorithms~\ref{alg:exact-dual-two-loss}
and~\ref{alg:exact-dual-three-loss}, respectively. For larger numbers of loss
terms, the dual subproblem is solved using a generic
Frank--Wolfe procedure. The resulting aggregate direction is then used to
recover the primal direction and the final scaled update direction, exactly as
in Algorithm~\ref{alg:dual-fw}, before applying Adam moments to the update.

\section{Experimental details}
\label{app:experiment_details}

\subsection{Software and hardware environments}
\label{app:software_hardware}

We conduct all experiments in a conda environment with Python 3.11, PyTorch 2.7.0+cu128, and CUDA 12.8. Our experiments are run on two servers: a Rocky Linux 9.7 server with 2 Intel Xeon Silver 4210R CPUs and 4 NVIDIA GeForce RTX 3090 24 GB GPUs, and an Ubuntu 24.04.4 LTS server with 2 Intel Xeon Silver 4210R CPUs and 4 NVIDIA GeForce RTX 2080 Ti 11 GB GPUs. Each complete experiment is run entirely on a single server to ensure fair comparison across methods.

\subsection{Details for section~\ref{sec:forward_experiments}}
\label{app:forward_details}

\paragraph{Benchmark equations}
The Helmholtz problem is defined by
\begin{align*}
& \Delta u(x,y) + k^2 u(x,y) = f(x,y), \qquad (x,y) \in \Omega, \\
& u(x,y) = 0, \qquad (x,y) \in \partial\Omega, \\
& \Omega = [-1,1]\times[-1,1].
\end{align*}
We use the exact solution
\[
u^\star(x,y) = \sin(a_1\pi x)\sin(a_2\pi y),
\]
which induces the forcing term
\[
f(x,y) = \bigl(k^2 - (a_1^2+a_2^2)\pi^2\bigr)\sin(a_1\pi x)\sin(a_2\pi y).
\]
In our experiments, we set \(k=1\), \(a_1=1\), and \(a_2=4\).

The 2D Kovasznay benchmark is governed by the steady incompressible Navier--Stokes equations
\begin{align*}
& u u_x + v u_y + p_x - \nu (u_{xx}+u_{yy}) = 0, \qquad (x,y)\in\Omega, \\
& u v_x + v v_y + p_y - \nu (v_{xx}+v_{yy}) = 0, \qquad (x,y)\in\Omega, \\
& u_x + v_y = 0, \qquad (x,y)\in\Omega,
\end{align*}
where
\[
\Omega = [-0.5,1]\times[-0.5,1.5].
\]
Dirichlet boundary conditions are imposed from the exact solution
\begin{align*}
& u^*(x,y) = 1 - e^{\lambda x}\cos(2\pi y), \\
& v^*(x,y) = \frac{\lambda}{2\pi}e^{\lambda x}\sin(2\pi y), \\
& p^*(x,y) = \frac{1}{2}\bigl(1-e^{2\lambda x}\bigr),
\end{align*}
with
\[
\lambda = \frac{1}{2\nu} - \sqrt{\frac{1}{4\nu^2}+4\pi^2}.
\]
We set \(\nu = 1/40\).

The viscous Burgers benchmark is defined by
\begin{align*}
& u_t(x,t) + u(x,t)u_x(x,t) - \nu u_{xx}(x,t) = 0, \qquad (x,t)\in[0,1]\times\Omega, \\
& u(x,0) = -\sin(\pi x), \qquad x\in\Omega, \\
& u(-1,t)=u(1,t) = 0, \qquad t\in[0,1], \\
& \Omega = [-1,1].
\end{align*}
In our experiments, we use \(\nu = 0.01/\pi\). Since this benchmark does not admit a closed-form exact solution under the above initial and boundary conditions, we evaluate it using a high-resolution numerical reference solution generated offline.

The 2D varying-coefficient heat benchmark is given by
\begin{align*}
& u_t(x,y,t) - a(x,y)\bigl(u_{xx}(x,y,t)+u_{yy}(x,y,t)\bigr) = f(x,y,t),
\qquad (x,y,t)\in\Omega\times[0,5], \\
& u(x,y,t) = 0, \qquad (x,y,t)\in\partial\Omega\times[0,5], \\
& u(x,y,0) = 0, \qquad (x,y)\in\Omega, \\
& \Omega = [0,1]\times[0,1],
\end{align*}
where
\[
f(x,y,t) = 200\sin(\pi x)\sin(5\pi y)\sin(\pi t).
\]
Here \(a(x,y)\) denotes the spatially varying diffusion coefficient taken from the PINNacle benchmark dataset~\cite{hao2024pinnacle}. For evaluation, we use the corresponding reference solution provided in the benchmark as a COMSOL-exported solution field.

The Klein--Gordon benchmark is defined by
\begin{align*}
& u_{tt}(x,t) - u_{xx}(x,t) + u(x,t)^3 = f(x,t), \qquad (x,t)\in[0,1]\times\Omega, \\
& u(x,0) = x, \qquad x\in\Omega, \\
& u_t(x,0) = 0, \qquad x\in\Omega, \\
& u(x,t) = u^*(x,t), \qquad (x,t)\in\partial\Omega\times[0,1], \\
& \Omega = [0,1].
\end{align*}
We use the manufactured exact solution
\[
u^*(x,t) = x\cos(5\pi t) + (tx)^3,
\]
which yields
\[
f(x,t)
= -25\pi^2 x\cos(5\pi t) + 6tx^3 - 6t^3x
+ \bigl(x\cos(5\pi t) + (tx)^3\bigr)^3.
\]

\paragraph{Network architecture and training setup}
For all forward benchmark problems, we use a fully connected multilayer perceptron with three linear layers, hidden width 50, and hyperbolic tangent activations. All weights are initialized with Xavier normal initialization~\cite{glorot2010understanding}, and all biases are initialized to zero. Each model is trained for 50,000 optimization steps on a single GPU. Following the implementation used throughout experiments, all methods adopt Adam~\cite{kingma2014adam} as the base optimizer with fixed parameters $\beta_1 = 0.9$, $\beta_2 = 0.999$, and $\epsilon = 10^{-8}$, and no learning-rate decay is used.

\paragraph{Sampling strategy}
For all forward benchmark problems, training points are resampled at every optimization step. Interior collocation points are drawn by uniform random sampling over the corresponding spatial or spatio-temporal domain. For boundary constraints, points are sampled uniformly on each boundary component; for the rectangular 2D domains, the boundary samples are distributed approximately evenly across the four edges, while for the 1D time-dependent problems they are sampled on the left and right spatial boundaries with uniformly sampled time coordinates. For initial conditions, the time coordinate is fixed at \(t=0\) and the spatial coordinates are sampled uniformly over the spatial domain. In the experiments, we use \(N_r=5120\) and \(N_b=256\) for the two steady-state problems, and \(N_r=2560\), \(N_b=256\), and \(N_i=256\) for the three time-dependent problems.

\paragraph{Optimization hyperparameters}
To account for differences in optimization behavior across methods, we select the learning rate separately for each method and benchmark from the set $\{10^{-3},\,3\times 10^{-4}\}$. Among the two candidates, we use the one that yields the better final relative $L^2$ error. The selected learning rates are summarized in Table~\ref{tab:forward_selected_lrs}. For LRA, we set $\alpha=0.1$ and update the loss weights every 10 steps. For ReLoBRaLo, we use $\alpha=0.95$, $\beta=0.99$, $\tau=1.0$, and $\epsilon=10^{-8}$. For MGDA, we use a Frank--Wolfe subsolver with maximum iteration 20 and tolerance $10^{-6}$. For our method, we set the stopping threshold to $10^{-6}$. 

\begin{table*}[t]
\centering
\small
\begin{tabular}{lccccc}
\hline
Method & Helmholtz & Kovasznay & Burgers & Heat 2D VC & Klein--Gordon \\
\hline
Adam      & \(10^{-3}\)  & \(10^{-3}\)  & \(10^{-3}\) & \(10^{-3}\)  & \(10^{-3}\)  \\
LRA       & \(3\times 10^{-4}\)  & \(10^{-3}\)  & \(10^{-3}\) & \(3\times 10^{-4}\)  & \(10^{-3}\)  \\
ReLoBRaLo & \(10^{-3}\)  & \(10^{-3}\)  & \(10^{-3}\) & \(10^{-3}\)  & \(3\times 10^{-4}\)  \\
MGDA      & \(10^{-3}\)  & \(10^{-3}\)  & \(10^{-3}\) & \(10^{-3}\)  & \(10^{-3}\)  \\
PCGrad    & \(10^{-3}\)  & \(10^{-3}\)  & \(10^{-3}\) & \(3\times 10^{-4}\)  & \(3\times 10^{-4}\)  \\
IMTL-G    & \(10^{-3}\) & \(3\times 10^{-4}\)  & \(3\times 10^{-4}\) & \(10^{-3}\)  & \(3\times 10^{-4}\)  \\
DCGD      & \(3\times 10^{-4}\)  & \(3\times 10^{-4}\)  & -- & -- & -- \\
ConFIG    & -- & -- & \(10^{-3}\) & \(10^{-3}\)  & \(3\times 10^{-4}\)  \\
HARMONIC  & \(3\times 10^{-4}\)  & \(3\times 10^{-4}\) & \(10^{-3}\) & \(10^{-3}\)  & \(3\times 10^{-4}\)  \\
Ours      & \(3\times 10^{-4}\)  & \(10^{-3}\)  & \(10^{-3}\) & \(10^{-3}\)  & \(3\times 10^{-4}\)  \\
\hline
\end{tabular}
\caption{Learning rates selected for the final reported runs in Section~\ref{sec:forward_experiments}. For each method and benchmark, the selection is made from \(\{10^{-3},\,3\times 10^{-4}\}\).}
\label{tab:forward_selected_lrs}
\end{table*}

\paragraph{Implementation detail for our method}
The main text presents our method using a generic dual subproblem solver. 
In the forward experiments in Section~\ref{sec:forward_experiments}, which use the
\(\ell_2\) setting, we use the exact variant of our method for the two-loss and
three-loss cases. In particular, for all reported forward runs of our method,
we use the implementation described in Appendix~\ref{app:algorithms}, which
solves the dual problem exactly when \(m=2\) or \(m=3\).

\paragraph{Evaluation metric}
We report the relative $L^2$ error as the evaluation metric. Given predictions $\hat{u}(x_i)$ and reference values $u_{\mathrm{ref}}(x_i)$ evaluated at $N$ points, we compute
\[
\frac{\left(\sum_{i=1}^N \left\|\hat{u}(x_i)-u_{\mathrm{ref}}(x_i)\right\|_2^2\right)^{1/2}}
{\left(\sum_{i=1}^N \left\|u_{\mathrm{ref}}(x_i)\right\|_2^2\right)^{1/2}}.
\]
For scalar-valued problems, the inner $\ell_2$ norm reduces to the absolute value. For the Helmholtz equation, 2D Kovasznay flow, and Klein--Gordon equation, $u_{\mathrm{ref}}$ is given by the corresponding analytic or manufactured solution. For Burgers' equation, $u_{\mathrm{ref}}$ is the offline high-resolution numerical reference described above. For the 2D varying-coefficient heat equation, $u_{\mathrm{ref}}$ is the benchmark reference solution field taken from the PINNacle dataset.

\subsection{Details for section~\ref{sec:inverse_experiments}}
\label{app:inverse_details}

\paragraph{Inverse problem setup}
We consider an inverse diffusion problem in which the solution \(u(x,y,t)\) and the spatially varying diffusion coefficient \(a(x,y)\) are learned simultaneously. The governing equation is
\begin{align*}
& u_t(x,y,t) - \nabla \cdot \bigl(a(x,y)\nabla u(x,y,t)\bigr) = f(x,y,t),
\qquad (x,y,t)\in\Omega\times[0,1], \\
& \Omega = [-1,1]\times[-1,1].
\end{align*}
We construct the benchmark from the reference fields
\[
u^\star(x,y,t) = e^{-t}\sin(\pi x)\sin(\pi y), \qquad
a^\star(x,y) = 2 + \sin(\pi x)\sin(\pi y),
\]
and choose \(f(x,y,t)\) accordingly so that \((u^\star, a^\star)\) satisfies the PDE.

\paragraph{Observation data and data settings}
To define the inverse problem, we use noisy pointwise observations of \(u\) together with boundary supervision for \(a\). The observation locations are drawn from a fixed pool of candidate points and then subsampled without replacement. More precisely, given observation points \(\{(x_j,y_j,t_j)\}_{j=1}^{N_{\mathrm{obs}}}\), we construct the observed values as
\[
\tilde{u}_j = u^\star(x_j,y_j,t_j) + \varepsilon_j, \qquad \varepsilon_j \sim \mathcal{N}(0,\sigma^2).
\]
The resulting noisy observation set is fixed throughout training. In addition, we impose coefficient boundary supervision by matching \(a(x,y)\) to \(a^\star(x,y)\) on \(\partial\Omega \times [0,1]\).

We consider three data settings in Section~\ref{sec:inverse_experiments}: \emph{Standard} with \(N_{\mathrm{obs}}=2500\) and \(\sigma=0.1\), \emph{Higher noise} with \(N_{\mathrm{obs}}=2500\) and \(\sigma=0.2\), and \emph{Fewer data} with \(N_{\mathrm{obs}}=1250\) and \(\sigma=0.1\).

\paragraph{Network architecture and training setup}
For the inverse heat problem, we use a parallel fully connected network with two separate branches, one for the solution field \(u(x,y,t)\) and the other for the coefficient field \(a(x,y)\). Each branch is implemented as a multilayer perceptron with five linear layers, hidden width \(50\), and hyperbolic tangent activations. To reflect the fact that the coefficient is time-independent, the \(a\)-branch takes only the spatial coordinates as effective inputs, while the time coordinate is masked out. All weights are initialized with Xavier normal initialization, and all biases are initialized to zero. Each model is trained for 30,000 optimization steps.

\paragraph{Sampling strategy}
At each optimization step, the collocation points and the coefficient-boundary points are resampled. The collocation points are drawn by uniform random sampling over \(\Omega \times [0,1]\), while the boundary points for \(a\) are sampled uniformly on the four spatial boundary components of \(\partial\Omega \times [0,1]\), with the samples distributed approximately evenly across the four edges. In contrast, the noisy observation set for \(u\) is generated once before training and then kept fixed throughout optimization. In the final experiments, we use \(N_r=5120\) collocation points and \(N_{b,a}=1024\) coefficient-boundary points per step. The number of observation points is \(N_{\mathrm{obs}}=2500\) for the Standard and Higher-noise settings, and \(N_{\mathrm{obs}}=1250\) for the Fewer-data setting.

\paragraph{Optimization hyperparameters}
The optimization hyperparameters are chosen in the same manner as in the forward experiments. In particular, for each method and data setting, we select the learning rate from \(\{10^{-3},\,3\times 10^{-4}\}\) and use the one that yields the better final relative \(L^2\) error. The selected learning rates are summarized in Table~\ref{tab:inverse_selected_lrs}. Method-specific hyperparameters are the same as those described above for the forward experiments.

\begin{table}[t]
\centering
\small
\begin{tabular}{lccc}
\hline
Method & Standard & Higher noise & Fewer data \\
\hline
Adam      & \(3\times 10^{-4}\)  & \(3\times 10^{-4}\)  & \(3\times 10^{-4}\)   \\
LRA       & \(3\times 10^{-4}\)  & \(3\times 10^{-4}\)  & \(10^{-3}\)   \\
ReLoBRaLo & \(10^{-3}\)  & \(10^{-3}\)  & \(10^{-3}\)  \\
MGDA      & \(3\times 10^{-4}\)  & \(3\times 10^{-4}\)  & \(3\times 10^{-4}\)  \\
PCGrad    & \(3\times 10^{-4}\)  & \(3\times 10^{-4}\)  & \(3\times 10^{-4}\)  \\
IMTL-G    & \(10^{-3}\)  & \(10^{-3}\)  & \(10^{-3}\)  \\
ConFIG    & \(10^{-3}\)  & \(10^{-3}\)  & \(10^{-3}\)  \\
HARMONIC  & \(10^{-3}\)  & \(10^{-3}\)  & \(10^{-3}\)  \\
Ours      & \(10^{-3}\)  & \(10^{-3}\)  & \(10^{-3}\)  \\
\hline
\end{tabular}
\caption{Learning rates selected for the final reported runs in Section~\ref{sec:inverse_experiments}. For each method and data setting, the selection is made from \(\{10^{-3},\,3\times 10^{-4}\}\).}
\label{tab:inverse_selected_lrs}
\end{table}

\paragraph{Evaluation metric}
As in the forward experiments, we use the relative \(L^2\) error as the evaluation metric. For the inverse problem, however, the reported error is computed on the recovered coefficient field \(a(x,y)\) rather than on the solution field \(u(x,y,t)\).

\section{Additional experimental results}

\subsection{Ablation Study on the Norm Parameter}
\label{app:ablation-p}

We conduct an ablation study on the norm parameter \(p\) in the proposed
\(\ell_p\)-based formulation. The experiment is performed on the Burgers forward
benchmark using the same setting as in the forward experiments, with \(2560\)
residual points, \(256\) boundary points, and \(256\) initial points. We compare
\(p\in\{1.5,2,3,4\}\), and each setting is repeated over three independent runs.
Table~\ref{tab:ablation_p_burgers} reports the mean and standard deviation.

Overall, the choices of \(p\) other than \(p=2\) do not improve performance on
this benchmark and generally lead to worse results.

\begin{table}[t]
\centering
\small
\begin{tabular}{lc}
\hline
Norm parameter \(p\) & Relative \(L_2\) error \\
\hline
\(1.5\) & 0.1317 (0.0522)  \\
\(2\)   & 0.0599 (0.0042)  \\
\(3\)   & 0.0966 (0.0238)  \\
\(4\)   & 0.1628 (0.0069)  \\
\hline
\end{tabular}
\caption{Ablation study on the norm parameter \(p\) for the Burgers forward benchmark. Each result is reported as mean and standard deviation over three independent runs.}
\label{tab:ablation_p_burgers}
\end{table}

\subsection{Computational Cost}
\label{app:computational-cost}

To compare the practical computational cost of the competing methods, we record the mean wall-clock time required to complete the full training run for each setting.
The forward benchmark problems are run for \(50{,}000\) training steps, whereas the inverse benchmark settings are run for \(30{,}000\) training steps. The measured runtimes, reported in seconds, are summarized in
Tables~\ref{tab:forward_computational_cost}
and~\ref{tab:inverse_computational_cost}, respectively.

\begin{table*}[t]
\centering
\small
\begin{tabular}{lccccc}
\hline
Method & Helmholtz & Kovasznay & Burgers & Heat 2D VC & Klein--Gordon \\
\hline
Adam      & 268.07  & 508.71  & 293.77  & 351.97  & 442.28 \\
LRA       & 283.39  & 525.18  & 320.22  & 334.14  & 429.72  \\
ReLoBRaLo & 323.73  & 526.12  & 361.85  & 372.15  & 473.33  \\
MGDA      & 288.43  & 530.61  & 616.87  & 408.62  & 574.87  \\
PCGrad    & 290.90  & 513.03  & 361.78  & 395.85  & 461.49  \\
IMTL-G    & 302.55  & 525.45  & 338.28  & 371.99  & 477.12  \\
DCGD      & 282.54  & 567.91 & -- & -- & -- \\
ConFIG    & -- & -- & 342.06  & 387.18  & 463.90  \\
HARMONIC  & 295.28  & 578.07  & 331.61  & 363.10 & 534.58  \\
Ours      & 311.12  & 540.58  & 366.63  & 405.21 & 501.02 \\
\hline
\end{tabular}
\caption{Mean wall-clock time required to complete \(50{,}000\) training steps for each forward benchmark problem.}
\label{tab:forward_computational_cost}
\end{table*}

\begin{table}[t]
\centering
\small
\begin{tabular}{lccc}
\hline
Method & Standard & Higher noise & Fewer data \\
\hline
Adam      & 642.93  & 574.93  & 587.18  \\
LRA       & 632.66  & 578.48  & 597.58  \\
ReLoBRaLo & 614.49  & 607.36  & 632.04  \\
MGDA      & 649.07  & 676.26  & 669.19  \\
PCGrad    & 647.41  & 610.27  & 626.52  \\
IMTL-G    & 612.95  & 601.61  & 596.25  \\
ConFIG    & 633.47  & 613.84  & 655.04  \\
HARMONIC  & 637.30  & 605.30  & 609.60  \\
Ours      & 676.37  & 648.24  & 624.63  \\
\hline
\end{tabular}
\caption{Mean wall-clock time required to complete \(30{,}000\) training steps for each inverse benchmark setting.}
\label{tab:inverse_computational_cost}
\end{table}

\end{document}